\documentclass[man,natbib,nofontenc,floatsintext]{apa6}
\usepackage[table]{xcolor}
\usepackage{graphicx}
\usepackage{amssymb,amsfonts,amsmath}
\usepackage{natbib} %[numbers,sort&compress]
\usepackage{setspace}
\usepackage[font=doublespacing]{caption}
\usepackage[T1]{fontenc}
\usepackage{multirow}
\usepackage{url}
\usepackage{tipa}

\usepackage{subcaption} 
\usepackage{makecell}

\newcommand\eg{\textit{e.g.,~}}
\newcommand\ie{\textit{i.e.,~}}

\title{Evaluating Models of Robust Word Recognition with Serial Reproduction}
\shorttitle{ROBUST WORD RECOGNITION}

\threeauthors{Stephan C. Meylan}{Sathvik Nair}{Thomas L. Griffiths}
\threeaffiliations{Department of Brain and Cognitive Sciences, Massachusetts Institute of Technology}{Department of Psychology, University of California, Berkeley}{Department of Psychology, Princeton University}

\abstract{
Spoken communication occurs in a ``noisy channel'' characterized by high levels of environmental noise, variability within and between speakers, and lexical and syntactic ambiguity.
Given these properties of the received linguistic input, robust spoken word recognition---and language processing more generally---relies heavily on listeners' prior knowledge to evaluate whether candidate interpretations of that input are more or less likely.
Here we compare several broad-coverage probabilistic generative language models in their ability to capture human linguistic expectations.
Serial reproduction, an experimental paradigm where spoken utterances are reproduced by successive participants similar to the children's game of ``Telephone,'' is used to elicit a sample that reflects the linguistic expectations of English-speaking adults.
When we evaluate a suite of probabilistic generative language models against the yielded chains of utterances, 
we find that those models that make use of abstract representations of preceding linguistic context (\ie phrase structure) best predict the changes made by people in the course of serial reproduction.
A logistic regression model predicting \textit{which} words in an utterance are most likely to be lost or changed in the course of spoken transmission corroborates this result.
We interpret these findings in light of research highlighting the interaction of memory-based constraints and representations in language processing.}
\keywords{spoken word recognition; sentence processing; generative language models; noisy-channel communication; iterated learning; serial reproduction}

\authornote{
\noindent We thank the Computational Cognitive Science Lab at U.C. Berkeley for valuable feedback.
This material is based on work supported by the National Science Foundation (Graduate Research Fellowship to S.C.M. under Grant No. DGE-1106400), the U.S. Air Force Office of Scientific Research (FA9550-13-1-0170 to T.L.G), and the Defense Advanced Research Projects Agency (Next Generation Social Sciences to T.L.G).

\noindent Address all correspondence to Stephan C. Meylan. E-mail: \texttt{smeylan@mit.edu}.}

\begin{document}

\maketitle

% N1.1. Linguistic communication occurs in a noisy channel; linguistic expectations are crucial
Spoken communication occurs in a ``noisy channel'' \citep{shannon1948, shannon1951}:
To overcome high levels of environmental noise, variation within and between speakers, and lexical and syntactic ambiguity, listeners must make extensive use of \textit{linguistic expectations}, or knowledge of what speakers are likely to say \citep{gibsonBergenPiantadosi2013}.
This requires the integration of many sources of information, including knowledge of word frequencies \citep{howes1957, hudson1985lexical, yap2012individual}, probable word sequences \citep{millerEtAl1951, smithLevy2013}, sub-word phonotactic regularities \citep{vitevitchLuce1999, storkel2006differentiating}, the plausibility of syntactic relationships \citep{altmannKamide1999, kamide2003time}, cues from the visual scene \citep{tanenhaus1995}, and pragmatic expectations \citep{rohdeEttlinger2012}.
More broadly, people are able to adjudicate between various candidate interpretations of auditory input in light of the specific discourse context, and bring considerable ``general world knowledge'' (e.g., knowledge of intuitive physics, properties of people and objects) to the task of natural language understanding \citep{levesque2011winograd}.

% N1.2. Introduce PGLMs: testable, quantitative; architectural distinctions; open qs
One particularly promising avenue of inquiry has been the development of statistical models that encode regularities in language structure in terms of probability distributions over sequences of words, or \textit{probabilistic generative language models} (PGLMs).
These models are distinguished in their ability to provide precise quantitative predictions for a variety of behavioral observables such as reading time and word naming latencies \citep{roarkEtAl2009, smithLevy2013, goodkindBicknell2018}. 
These models are further distinguished in their ability to reflect human-relevant psycholinguistic hypotheses in their architecture, for example whether syntactic structure is  symbolically represented or implicit \citep{frankBod2011, fossumLevy2012}.
However, as we argue below, there are a number of substantive challenges and limitations in the way these models are currently evaluated, especially in the auditory modality. 
This leaves important open questions regarding which of these models best capture human expectations in spoken word recognition in sentential contexts. 

\begin{figure}[t]
\begin{center}
\includegraphics[width=\textwidth]{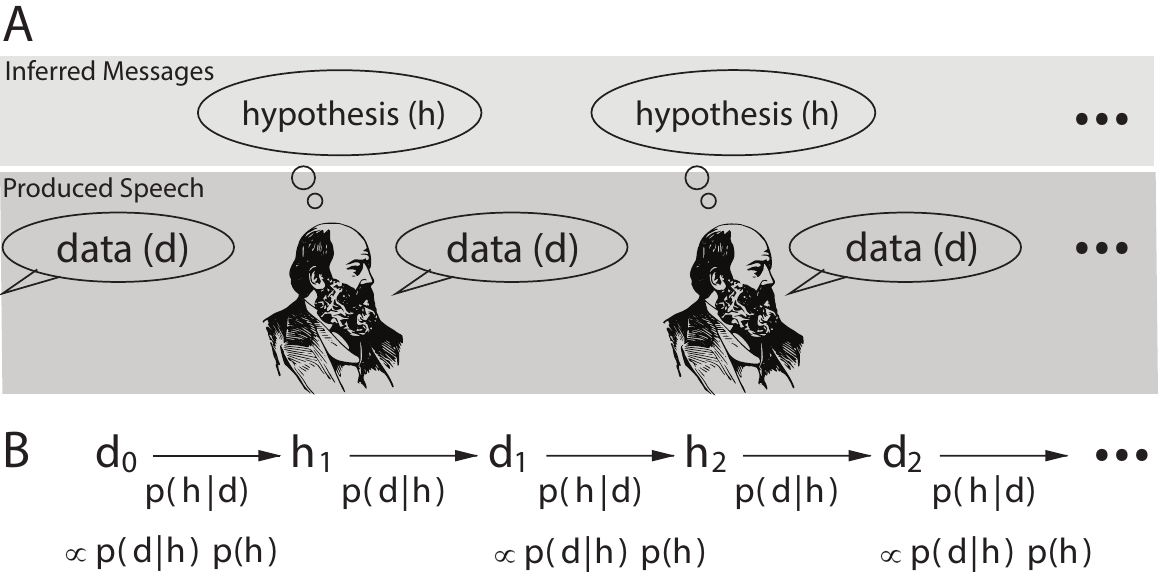}
\end{center}
\vspace{-5mm}
\caption{\label{fig:serialReproduction} 
A: Serial reproduction. Each participant chooses a hypothesis regarding what they heard, and on the basis of that hypothesis produces data for the next listener. B: Under the Bayesian model of the telephone task presented here, participants assign posterior probability $p(h|d)$ to hypotheses regarding what they heard  proportional to the product of the likelihood, $p(d|h)$, which depends on the current auditory input, and the prior, $p(h)$, which does not. 
}
\end{figure}

%N1.4: Outline the methods
The behavioral experiment presented here specifically targets a gap in methods for estimating people's linguistic expectations in the audio modality, \textit{i.e.}, what people expect other people to say.
In the current work, we develop a method to sample from the linguistic expectations used by people in a naturalistic language task using the technique of \textit{serial reproduction}  \citep{bartlett1932, xuGriffiths2010}, in this case a large-scale web-based game of ``Telephone''.
A succession of participants reproduces each initial sentence, such that the utterance produced by each participant is used as input to the next (Figure \ref{fig:serialReproduction}).
Similar to the related technique of \textit{iterated learning} \citep{kirby2001}, this technique of serial reproduction can be shown to converge to participants' prior expectations given a sufficient number of iterations (repetitions, with changes introduced by participants), and as long as certain conditions are met \citep{griffithsKalish2007}.
Even with fewer iterations than required for convergence to the stationary distribution, the yielded samples are still useful in that they reflect an approach to the distribution of interest.

%FIGURE: SERIAL Reproduction 

%N1.5: results preview
To preview our results: the changes people make to sentences in the Telephone game are best explained by probabilistic generative language models that make use of abstract symbolic representations of preceding context.
A token-level analysis of which individual words are successfully transmitted from one participant to another provides converging evidence for this result.
More generally, we introduce a method to elicit samples from people's expectations for receptive language tasks in the audio modality, and show how that method can be used to evaluate language models in their ability to explain key human linguistic behaviors.

\section{Theoretical Background}
%N2: Theoretical background

% Not sure how to relate the theoretical background to the 

%N2.1 How do we approximate language model? Levy: bottleneck allows for a dialectic among models

\begin{figure}[t]
\begin{center}
\includegraphics[width=5.5in]{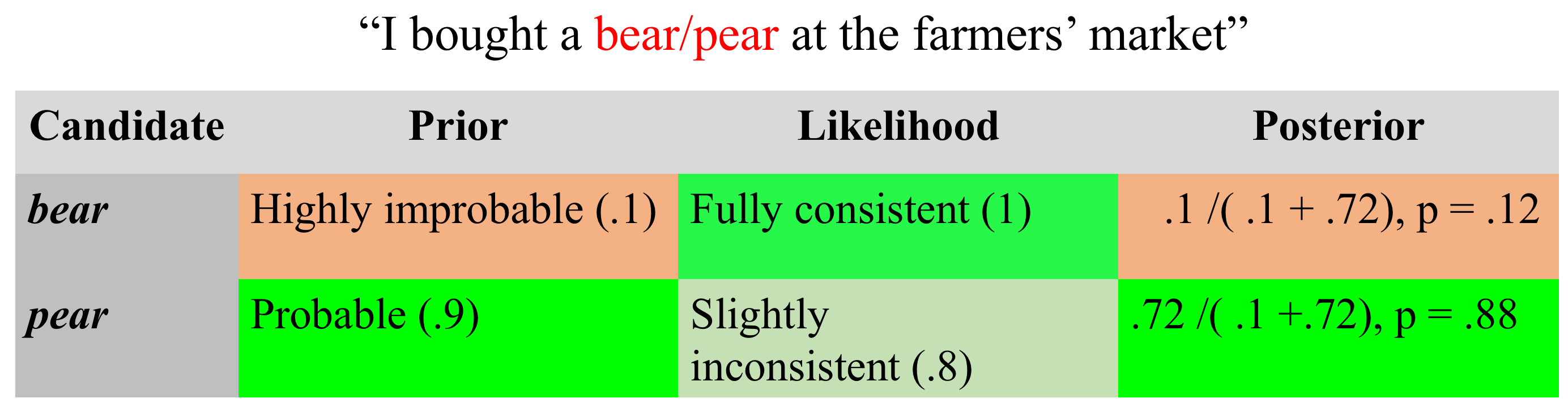}
\end{center}
\vspace{-5mm}
\caption{\label{fig:bearPear} 
Example of the contributions of the likelihood and prior in spoken word recognition in a sentential context. The phonemes /b/ and /p/ are largely distinguished by a single dimension of difference, their voice onset time (VOT), and are relatively easily confused.
Using broader language knowledge, a listener can recognize a speaker's likely intended meaning when data is ambiguous or noisy.
In this toy example, the prior probability of \textit{pear} in this context overwhelms the evidence for \textit{bear} provided by the received auditory signal.
}
\end{figure}

In a landmark study, \citet{lucePisoni1998} showed that the recognition of isolated words embedded in noise can be modeled as a competitive process, combining evidence from the received auditory signal with each word's probability (\ie relative frequency) in a corpus.
Subsequent work has formalized this process of competition among words within an explicitly Bayesian framework, and extended it to the recognition of words in sentential contexts \citep{norrisMcQueen2008}.
It is now widely accepted that linguistic expectations---probabilistic knowledge regarding what is more or less likely to be said---make the process of word recognition substantially more robust than it would be as a purely data-driven, bottom-up process \citep{norrisEtAl2016}.
Figure \ref{fig:bearPear} provides a toy example of how a listener might overcome perceptual noise to arrive at a speaker's intended message.
While a listener may receive auditory input more consistent with \textit{bear} than \textit{pear} after the preceding context \textit{I bought a}, the probabilities of the two candidates as continuations given the preceding context would suggest that the latter is a much more likely interpretation.
In this way, a listener can overcome noise and ambiguity and successfully recover a speaker's message using their knowledge of language. 

%FIGURE: bear/pear toy example of Bayesian inference 

A critical question at Marr's computational level of analysis \citep{marr1982} is what sort of information sources might be combined to accomplish the task of word recognition, independent of the precise timecourse or computational tractability. 
\citet{lucePisoni1998} use word probability---normalized frequency---as an example of basic word-level knowledge that listeners have access to before encountering auditory input.
This simple model would, however, often lead to incorrect predictions for the bear/pear example in Figure \ref{fig:bearPear}, because \textit{bear} is more frequent than \textit{pear} in many linguistic corpora.
Rather, the plausibility of the two candidates reflects more detailed world knowledge, for example what sort of things might be bought and sold at a farmer's market.
In the absence of models that encode this sort of world information directly, increasingly sophisticated models of linguistic structure, or probabilistic generative language models (PGLMs), have been used to approximate people's expectations.

Following on the work of \citet{hale2001}, \citet{levy2008} tested how the probabilistic linguistic expectations encoded by such a model predicts sentence processing difficulty, as revealed by measures of reading time.
This work specifically advanced the concept of a ``causal bottleneck:'' the expectedness of a word determines its processing difficulty, and this expectedness may reflect a broad range of linguistic and non-linguistic knowledge.
The negative log probability of that upcoming word under a listener's expectations, or \textit{surprisal} (see also \citealp{itti2009bayesian} for a related formulation in the visual domain), provides a succinct measure of a word's expectedness.
\citet{levy2008} further demonstrated that surprisal estimates derived from a PGLM can be used to approximate people's rich implicit knowledge of linguistic regularities, and can provide significant explanatory power above and beyond theories of processing difficulty that focus on memory constraints alone.

%N2.2 Computational models from speech recognition: need prior context

Though \citet{levy2008} used a specific probabilistic generative model (the Earley parser of \citealp{hale2001}), the theory of a causal bottleneck posits a broader relationship between surprisal and processing difficulty.
Specifically, more sophisticated models---those capable of capturing additional information sources available to people---should be expected to produce surprisal estimates more strongly correlated with observed processing difficulty. 
Thus while these models entail strong simplifying assumptions regarding the knowledge of language structure available to people (let alone knowledge of language structure available to linguists), they can nonetheless be used to derive expectations from large corpora or other large-scale datasets that are strongly predictive of human linguistic behavior, such as response times in word naming tasks.
Subsequent work in psycholinguistics and cognitive science has demonstrated the utility of a variety of PGLMs in understanding human sentence processing, especially for reading \citep{dembergKeller2008, demberg2009, frankBod2011, fossumLevy2012, smith2011cloze, smithLevy2013, fineEtAl2013, futrell2017noisy, goodkindBicknell2018}.
% and shed considerable light on communicative constraints on lexicons \citep{piantadosiEtAl2011, mahowaldEtAl2013}.
However, the question of \textit{which} models best capture human linguistic expectations---using which information sources and via which representations---remains a central open question.

In addition to the utility of increasingly accurate statistical models of language structure for psychology and cognitive science, these same models are of critical importance for a broad range of speech-related engineering applications, where noise, ambiguity, and speaker variation have been well-known challenges since the 1940s (e.g., \citealp{shannon1948}).
Generative models of utterance structure have received extensive treatment in the fields of Computational Linguistics, Natural Language Processing, and Automatic Speech Recognition, where such models are collectively known as ``language models'' \citep{jurafskyMartin2009}.
Their probabilistic form allows them to be combined in principled ways with auditory information using Bayesian techniques, or approximations thereof (\eg \citealp{graves2006connectionist}).
A growing interest in commercial applications in recent years has resulted in a profusion of model architectures, especially ones taking advantage of large-scale deep artificial neural networks (e.g., \citealp{hannunEtAl2014, zarembaEtAl2014, jozefowicz2015empirical, dyerEtAl2016, gulordava2018colorless, petersEtAl2018, devlin2018bert}).
These neural network models have received significant attention in their ability to capture human linguistic phenomena above and beyond simpler structural models, though the exact nature of such remains an active area of investigation \citep{linzenEtAl2016, linzen2018distinct,  futrellEtAl2019, futrellLevy2019}.

%N2.3 Quick overview of models, with reference to the appendix

\section{Language Models}

The primary objective of this contribution is to evaluate a variety of broad-coverage probabilistic generative language models (PGLMs) in their ability to capture the linguistic expectations implicit in participants' behavior in a game of Telephone.
By ``language models,'' we follow the convention of the natural language processing literature to refer to probabilistic models that encode information regarding the distribution over possible continuations (the next word in the utterance) before encountering the relevant auditory input.
We thus exclude many ``language models'' in the broader sense that are concerned with the online dynamics of word recognition like the logogen model \citep{morton1969}, cohort model \citep{marslenWilson1987}, or TRACE model of word recognition \citep{mcclelland1986trace}.
In that modeling human performance in the Telephone game requires large vocabularies and a means of handling newly-encountered words, all language models here track expectations over 10,000 or more word types and have a method for handling out-of-vocabulary items.

\begin{table}[t]
\footnotesize
\caption{Overview of probabilistic language models used in the current work.\label{tab:model_overview}}
\begin{tabular}{llll}
Model                         & Type            & Training Data                            & Citation \\
\hline
OBWB / BIG LSTM               & LSTM            & One Billion Word Benchmark               & Jozefowicz et al. (2015) \\
BNC / Kneser-Ney Trigram      & Smoothed N-gram & British National Corpus                  & Chen and Goodman (1998)\\
BNC / Unigram                 & N-gram          & British National Corpus                  &  \\
DS / Kneser-Ney 5-gram        & Smoothed N-gram & Librivox, Fisher, Switchboard & Chen and Goodman (1998) \\
PTB + GTB / BLLIP  & PCFG            & Penn + Google TreeBanks        & Charniak and Johnson (2005) \\
PTB / BLLIP            & PCFG            & Penn TreeBank                           & Charniak and Johnson (2005)  \\
PTB / Roark Parser            & PCFG            & Penn TreeBank                           & Roark et al. (2009) \\
PTB / RNNLM                   & RNN             & Penn TreeBank                           & Mikolov et al. (2010) \\
PTB / Unigram                 & N-gram          & Penn TreeBank  &  \\
PTB / Good-Turing Trigram     & Smoothed N-gram & Penn TreeBank                           & Gale and Sampson (1995) \\
PTB / Good-Turing 5-gram      & Smoothed N-gram & Penn TreeBank                           & Gale and Sampson (1995) 
\end{tabular}
\end{table}

%!!! description of the major models
The language models evaluated here are outlined in Table \ref{tab:model_overview}.
They include $n$-gram models of varying orders, parsers that use probabilistic context-free grammars, and recurrent neural networks, including LSTMs.
Where possible, two versions of each model are evaluated: one fit/trained on the Penn TreeBank, and one with a large dataset known to yield competitive prediction results.
For additional details, refer to \textit{Technical Appendix 1: Language Models and Training Procedure}. 
Insofar as each model is the subject of many dissertations, conference papers, and journal articles in its own right, the overviews provided here are neither exhaustive nor formally rigorous; we refer readers to the appropriate publications for more details. 

% https://www.quora.com/What-perplexities-do-state-of-the-art-language-models-achieve
%!!! table here with an overview of the models

The PGLMs under study here vary in structural complexity from none at all (a unigram model, which may be considered a null model positing no structure at all) to distributions over parse trees, positing hierarchical syntactic relations, such as noun phrases and verb phrases, for all words in a sentence.
Language models with higher structural complexity bring the task of in-context spoken word recognition into contact with the task of \textit{sentence processing}: if people think that the utterances they hear will adhere to grammatical rules, their expectations about upcoming words will be quite different.
While people certainly use detailed representations of relationships between words in sentence understanding more broadly, it remains an open question whether this same information is used online in the process of identifying words in noisy linguistic input \citep{connineClifton1987, levyEtAl2009, mcQueenAndHuettig2012, lashEtAl2013}.

\subsection*{Abstract Structure in Preceding Context}

One of the central distinctions that we address here is whether there is evidence that people use \textit{categorical, hierarchical} syntactic representations in the service of prediction. 
Contrary to previous work finding evidence of extensive use of abstract representations in sentence processing, (e.g., \citealp{gibsonEtAl2005}), \citet{frankBod2011} found no additional predictive power for models that use explicit hierarchically-structured representations of preceding sentence context for predicting reading times in the Dundee corpus.
The models under comparison in their study included $n$-gram models, the Roark parser \citep{roark2001}, and echo state networks \citep{jaeger2001}, a kind of recurrent neural network architecture.
They further expand on this position and describe the prospects and implications of a non-hierarchical theory of language processing in \citep{frank2012hierarchical}.
By contrast, a replication and extension of \citet{frankBod2011} by \citet{fossumLevy2012} found that the use of improved lexical $n$-gram controls derived from a larger model and using a more sophisticated smoothing technique eliminated the performance differences between unlexicalized sequential and hierarchical models.
Further, they showed that a state-of-the-art \textit{lexicalized} hierarchical model from \citet{petrovKlein2007}---a model that tracks more granular relationships between words and grammatical categories---predicts reading times better yet. 
We expand on the contrast between lexicalized and unlexicalized models below.
\citet{vanSchijndel2015} also found evidence that models with hierarchical representations of syntactic context better predict reading times in the same corpus.

In the current work, we investigate the utility of abstract representations in the auditory domain, using a larger sample of language models.
We adopt a different typology of models than that used in the above works: we group recurrent neural network models---of which we include two more recent architectures---with models that infer parse trees.
While \citet{fossumLevy2012} and \citet{frankBod2011} separate those models that make full hierarchical representations of the phrase structure (``latent context in the form of syntactic trees'') vs. others, we note that recurrent neural networks, such as the echo state network used by \citet{frankBod2011}, capture significant higher-order linguistic regularities of a qualitatively similar type to the nonterminals in a phrase structure grammar.
An examination of the word embeddings from \citet{elman1990}, a simple recurrent neural network, show that the average activation pattern for a word reflects a combination of \textit{both} semantic and syntactic similarity, in that both of these are reflected in the lexical distributions in the surrounding context.
While this is admittedly far from a full hierarchical parse tree, the use of \textit{any} sort of abstract representation of context may have a stronger effect on the sort of expectations that are encoded in a model, rather than whether the representation is represented symbolically as hierarchical.
We evaluate support for this partition of models, as well as the performance of the language models in predicting the pattern of changes, using our collected serial reproduction data.
Following \citet{frank2018hierarchical}, we do not entertain the hypothesis that language processing and especially sentence processing is devoid of hierarchical phenomena (e.g., \citealp{altmannKamide1999}).
Rather, we focus our investigation on whether hierarchical or non-hierarchical models better account for the data collected from our task, specifically investigating the role of prediction in in-context word recognition.

%N2.4 problems with other ways to evaluate PGLMs
%	- Probability of a held-out test set (perplexity)		
%	- Reading time: Confound of errors vs. behavioral processing difficulties	
%	- Fit to behavioral observables
%		- Goodkind and Bicknell -- reading; spoken word recognition may have different characteristics 
%		Correlation between perplexity and prediction on behavioral observables
%		(this is a more reasonable way to do this than my in same approach)

As noted in the introduction, PGLMs are typically evaluated according to the probability that they assign to a test sample of language (in natural language processing) or in their ability to predict experimental observables pertaining to processing difficulty (in psycholinguistics).
Each of these evaluation methods has notable limitations.

Engineering-centric research in natural language processing typically evaluates generative language models in terms of the probability that they assign to a held-out dataset: the model that assigns a higher probability to a held-out dataset (or \textit{perplexity}, in which probability is normalized by its length in words) is the better model in the absence of confounding factors \citep{jurafskyMartin2009}.
This metric assigns the highest scores to those models that best capture the statistics of the corpora they are tested on.
Fitting language models to large naturalistic corpora produces excellent first-approximation, broad-coverage estimates of people's general-purpose language expectations.
However, people's expectations may deviate significantly from these corpus statistics: people do not expect newly encountered speakers to produce utterances following the statistical properties of corpora derived from books or news (\eg the Penn TreeBank).
Even the use of naturalistic corpora like Switchboard \citep{godfreyEtAl1992} or Santa Barbara \citep{duBoisEtAl2000} fails to address this problem, in that participants' expectations for the purposes of comprehension may deviate significantly from that of any known corpora, a point on which we elaborate below.
Likewise, eliciting a sample of unconditioned speech may not reflect a person's expectations of what \textit{others} are likely to say.

Regarding maximizing test sample probability, there is no guarantee that a corpus-derived test set has a high probability under human linguistic expectations. 
As rational agents, people should be expected to develop expectations that match the task of interpreting linguistic material under noisy conditions, but in the case of language it is very hard to know what will constitute future ``typical'' linguistic experience.
Composition in terms of written vs. auditory sources, topics, and speech registers (i.e., levels of formality) is undoubtedly subject to significant individual variation.
Furthermore, even with a perfect estimate of a person's experience with language thus far, there is no guarantee that linguistic material encountered in the future will have the same composition as that encountered in the past.
In such a case, a rational agent might be expected to allocate probability mass to as yet unobserved linguistic events to avoid overfitting on the basis of previous experience.
As such, we argue that \textit{no} corpus collected from natural sources should be expected to reflect people's linguistic expectations for the task of in-context word recognition.\footnote{We note, however, that language models fit on large corpora of naturalistic text are an excellent first approximation to people's linguistic expectations, and that they provide broad coverage for relatively little effort by comparison to the iterated learning method introduced here. We return to this point, as well as prospects for combining the two approaches, in the Discussion.}

Psycholinguistic studies, by contrast, have typically evaluated how well surprisal estimates derived from probabilistic language models predict measures of processing difficulty, such as reaction times in word naming, looking time or regressions in reading as measured by eye tracking, and neural measures such as the magnitude of event-related potentials.
We note a challenge in interpreting the relationship between surprisal and processing difficulty.
Unpredictable linguistic events should cause increased processing difficulty \textit{if and only if} they are correctly interpreted:  they may also simply be missed by listeners/readers and result in transmission errors (i.e., a listener could incorrectly interpret an unexpected word as a more probable alternative, without immediately incurring any increase in processing difficulty).
As such, behavioral observables constitute an imperfect record of human linguistic expectations.
Models should thus be evaluated with respect to how surprisal estimates predict measures of processing difficulty, as well as the kinds of transmission errors they predict.

\subsection{Characterizing Transmission Errors}

A substantial body of work in signal processing has evaluated the performance of automatic speech recognition systems (ASR) on conversational speech \citep{lippmann1997}.
While most of this work has focused on engineering applications, some work has treated these systems as computational-level cognitive models of word recognition.
\citet{goldwaterEtAl2010}, for example, evaluated two contemporary ASR systems in their ability to recognize words from a standard dataset.
Consistent with the human results reviewed below, they found higher error rates for low probability words, with strong independent effects of unigram and trigram probabilities.
They also found that longer words have lower error rates, that  words are more likely to be mis-recognized at the beginnings of utterances, and that the model was particularly prone to transmission failures among ``doubly confusable pairs:'' pairs of words with similar acoustic profiles that appear in similar lexical contexts.
More recently, \citet{xiongEtAl2018} found that neural network-based  ASR systems make similar errors to people, though many of the remaining differences can be traced to how these models handle backchanneling (e.g. ``mm-hmm'') and disfluencies (``uh'',``hmmm''). 
This work evaluating ASR systems has, however, systematically evaluated how transmission errors vary as a function of language model.

Word recognition errors have also been thoroughly explored in psycholinguistics and audiology.
Early work focused on establishing psychometric functions to relate signal-to-noise ratios of word presentation in noise to recognition accuracy  \citep{egan1948, frenchSteinberg1947}.
Subsequent work tested the effects of word frequency \citep{owens1961, savin1963, broadbent1967}, the interaction of word frequency and acoustic similarity \citep{lucePisoni1998}, and the importance of lexical neighborhoods (the number of phonetically similar words to a target word,  \citealp{treisman1978};  \citealp{luceEtAl1990}; \citealp{zieglerEtAl2003}) as determinants of word recognition accuracy.
While much of this work focuses on the recognition of individual words in isolation, a major thread explores the facilitatory effects of utterance context  \citep{millerEtAl1951, kalikowEtAl1977, connineEtAl1999, benichovEtAl2012}.
We note however that these studies have largely focused on recognition accuracy in specific sentence positions, usually with small sets of carefully  experimentally-controlled stimuli.
Further, the characterization of predictability has relied on small corpora, and predictability estimates have often been coded as categorical rather than continuous variables.
As noted by \citet{theunissenEtAl2009}, experiments investigating the transmission of entire utterances have rarely characterized the word-level changes.
In the current work, we adopt a methodology that allows us to track transmission errors among \textit{all} words in a large collection of utterances, and uses predictability estimates from sophisticated language models trained on large datasets.

One challenge for word recognition research is that there is a rich correlational structure among word features: shorter words are more frequent, have more phonologically similar competitors, are learned earlier, and are more central in lexical networks \citep{vincentLamarreEtAl2016}. 
In the absence of controlling for these variables, this makes it difficult to interpret many of the above results.
More recent work addresses this challenge by including a larger set of predictors, explicitly characterizing the correlational structure among those predictors, and using linear and logistic mixed-effects models to account for participant-level and item-level variability. 
Looking at this expanded set of predictors and using appropriate regression models, \citet{gahlStrand2016} reproduced the longstanding effect of lexical frequency (more frequent words are easier to recognize) and found that words with many phonological neighbors are harder to recognize.

More broadly, there are several reasons to question whether the effects observed in isolated word recognition---which has received the overwhelming balance attention in the literature thus far---will extend to the recognition of words in utterance contexts.
First, many candidate interpretations may be much less competitive when contextual cues are available to guide processing.
In isolated word recognition, participants may adapt to the circumstances of the task (e.g., expect that low-frequency and high-frequency words are equally likely in an experimental context), and thus diverge from expectations for in-context conversational speech.
Second,  many of these studies have investigated only a limited set of words,  for example only monosyllabic words or words without complex consonant clusters.
The current work thus provides an opportunity to test whether patterns observed in isolated word recognition experiments extend to the task of in-context speech recognition.

%N2.5: Serial reproduction as a solution? With some amount of tempering ? not that it yields the best new thing 

\subsection*{Serial Reproduction}

In the present study, we use the technique of \textit{serial reproduction} to investigate human in-context word recognition.
Intuitively, we start with a small test corpus and use serial reproduction to gradually change the properties of that corpus so that it better reflects people's linguistic expectations.
This transition in corpus properties emerges in the course of serial reproduction because participants' expectations are reflected in the changes they make between the utterance they hear and the utterance they produce at each iteration.

Information transmission by serial reproduction was first studied by Sir Frederic Bartlett, who tracked the evolution of stories and pictures recreated from memory after rapid presentation \citep{bartlett1932}.
More recent work has identified that the technique, like iterated learning \citep{kirby2001}, can be used to experimentally reveal inductive biases, or reasons that people would favor one hypothesis over another independent of observed data in inferential tasks \citep{mitchell1997}. 
If participants (reproducers for serial reproduction, learners for iterated learning) use Bayesian inference to infer the posterior distribution over hypotheses, and then draw from that distribution in production, then their output at each sequential generation reflects a combination of observed data and participants' inductive biases.
Over time, the distribution implicit in the output data comes to reflect participants' inductive biases.
For a trial in the Telephone game, participants' hypotheses are the set of possible interpretations that they might provide for an utterance, in that the message they infer reflects both the auditory input and their expectations regarding language. 
%We model these expectations regarding language---the prior a listener uses when inferring a speaker's message--with a suite of probabilistic generative language models (PGLMs), which are described in greater detail below.   

Under certain assumptions, both kinds of transmission chains (iterated learning and serial reproduction) can be interpreted as a form of Gibbs sampling, a common technique in Markov chain Monte Carlo based parameter estimation \citep{gemanGeman1984}. 
The transmission of a linguistic utterance across a succession of participants can be interpreted as a Markov process; given certain assumptions, the resulting Markov chain is guaranteed to converge to its stationary distribution if run for a sufficient number of iterations.
A participant estimates the probability of the interpretation of an utterance $h$ given some observed data (i.e. an audio recording) $d$ following Bayesian inference (Figure \ref{fig:serialReproduction}).
In particular, following a process of sampling a hypothesis $h$ from the posterior, $p(h|d) \propto p(d|h)p(h)$ and sampling data from $p(d|h)$ means that the probability that a participant selects a hypothesis $h$ converges to their prior distribution $p(h)$ \citep{griffithsKalish2007}. 
This approach has been profitably used to reveal inductive biases in memory \citep{xuGriffiths2010},  category structure \citep{griffithsEtAl2008, sanbornEtAl2010}, and function learning \citep{griffithsKalish2008}.
In the domain of language specifically, iterated learning has been used to reveal biases relevant to language evolution \citep{griffithsKalish2007}.

We motivate the use of serial reproduction with language in the auditory modality by analogy to the function learning experiments of \citet{griffithsKalish2008}, which can be thought of as a game of ``Telephone'' in the space of mathematical functions.
In these experiments, participants saw a small number of points in a two-dimensional space drawn from an unknown---and not necessarily linear---function.
They were then asked to provide their best guess regarding the function which generated the observed points.
This function was then used to generate points for the next participant in the chain.
While different transmission chains started with radically different functions, including some of considerable complexity, all are reduced to simple, mostly positive linear functions in the course of reproduction.
In the face of limited, noisy data people tend towards their expectation for a simple linear function.
The case of language implicates a more complex and nuanced set of expectations---what people expect other people to say---but the basic experimental logic remains the same.\footnote{An intuitive interpretation of this process is that repetition entails reversion to the mean. In the case of linguistic expectations, this ``mean'' is of unknown character and the subject of broad theoretical interest.}

%Figure: Inductive bias in function learning -- don't use for brevity + copyright

%\begin{figure}[t]
%\begin{center}
%\includegraphics[width=5.5in]{figures/linear_iterated.png}
%\end{center}
%\vspace{-5mm}
%\caption{\label{fig:linear_iterated} 
%Example transmission chains from a function learning experiment.
%Columns (i-x) correspond to successive participants reproducing different initial functions, ($a_{i} \ldots d_{i}$).
%The resulting positive, linear functions ($a_{x} \ldots d_{x}$) provide evidence of an inductive bias for function learning.   
%Figure reproduced from \citet{griffithsKalish2008}.
%}
%\end{figure}

The process of serial reproduction can consequently be thought of as gradually changing a corpus so that it better reflects what people expect others to say.
To the degree that the initial set of utterances in that corpus is not representative of what people expect to hear, then the process of serial reproduction will introduce edits that change them to increasingly reflect the broader linguistic expectations of participants.
If, for example, the initial corpus contained an excessive number of low-frequency words pertaining to finance, participants should be expected to misinterpret these and replace them with words prototypical of normal conversational registers.
To our knowledge, no previous work has used this technique of serial reproduction to investigate language processing.
However, we note that the task of serial reproduction of isolated speech sounds was previously used to investigate whether repeated imitation of environmental noises can give rise to word-like forms \citep{edmiston2018repeated}. 

\section{Methods}

We use a web experiment to gather chains of audio recordings and transcriptions appropriate for answering our primary research questions.
This web experiment lets participants listen to and record audio, coordinates the flow of stimuli such that recordings from one participant can be used as stimuli for later participants, and ensures that the succession of recordings remains interpretable linguistic material, while avoiding explicit judgments of appropriateness by the experimenters.
For simplicity of exposition, we first describe how a participant would experience the task; then we describe the flow of data in the experiment, focusing on how the successive contributions of participants are used as stimuli for future participants.

\subsection*{Experimental Interface}

Upon accepting the experiment on Amazon Mechanical Turk, a participant is provided with a link to a web application that allows them to listen to and record responses to audio recordings of utterances.
Participants proceed through two practice trials that test their speakers and microphone. 
A participant is then introduced to the full trial format, where 1) they listen to a recording 2) they choose whether or not to flag the recording they heard as appropriate and interpretable 3) they record a response (i.e., their best guess of the content of the utterance they heard) 4) they decide whether to flag their \textit{own} recording (\eg in case of speech errors or excessive ambient noise) 5) they provide a written transcription of the utterance they recorded. 
After the participant submits a trial, the new recording and transcription are programmatically evaluated with a number of filters confirming that the audio recording is not blank, that it is of similar length to the previous utterance, and that it passes basic tests of consistency between recording and transcription using an automated speech recognition system.
The participant does two practice trials in this full trial format to confirm their understanding of the directions of the task; if they fail to pass the basic checks, they continue through additional practice trials.
For additional details regarding the web interface and the automated checks, refer to \textit{Technical Appendix 2: Experiment Interface}.

% FIGURE: screenshot of the experiment
\begin{figure}
\begin{center}
\includegraphics[width=6.5in]{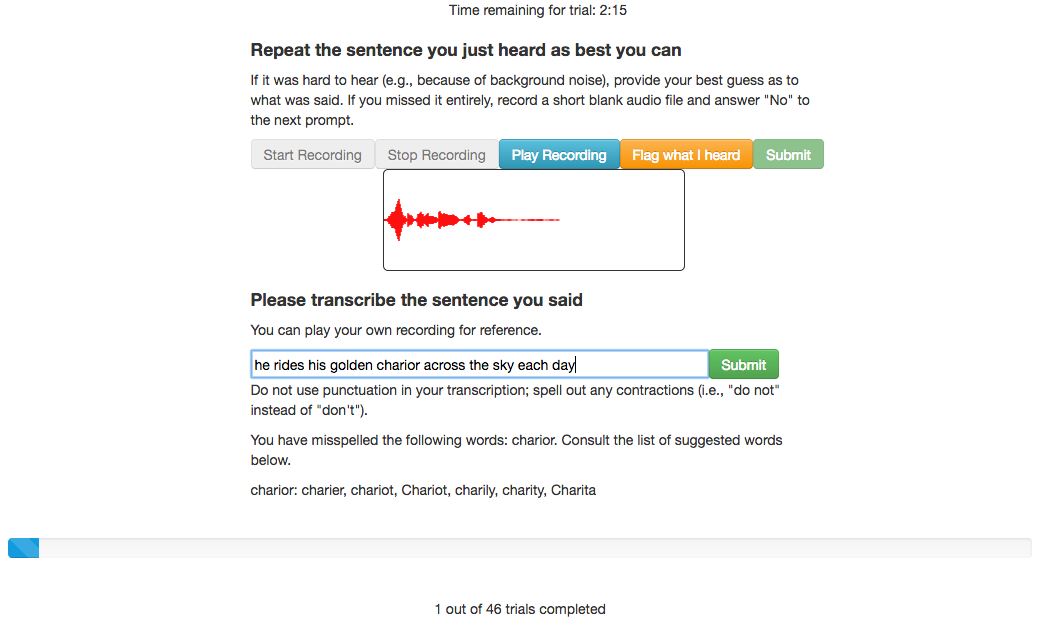}
\end{center}
\vspace{-5mm}
\caption{\label{fig:telephoneInterface} 
The browser-based audio recording and transcription interface used by participants.
This screenshot depicts the state of the application after the participant has listened to a recording by a previous participant, recorded a response, and is in the process of submitting a transcription for their contributed recording. In this case the participant has been flagged for misspelling a word.
}
\end{figure}

\subsection*{Stimuli and Experimental Design}

We select a set of 40 initial target sentences and 9 filler sentences from the TASA corpus \citep{zenoEtAl1995}  and the Brown Corpus \citep{kuceraFrancis1967}.
These sentences are chosen to provide maximal variation in probability under different language models for sentences of the same length. 
First we determined which sentence length (in terms of words and characters) was the most common in the corpus (10 words, 42 nonspace characters + 9 spaces).
All were evaluated as grammatical by the experimenters.
For this cohort of length-matched sentences, we then obtain their probabilities under unigram and trigram language models trained on the British National Corpus.
For the trigram model, we used modified Kneser-Ney smoothing \citep{chenGoodman1998} on transitions of order three (for further details regarding this smoothing scheme, see Language Models below).
For both unigram and trigram probabilities, we used the empirical distribution of probabilities for the yielded sentences to generate 20 5-percentile tranches. 
Then for each tranch, we iterated through sentences, rejecting all sentences phrased as questions, interpretable as inappropriate, or containing numerical characters, hyphenated words, or contractions.
This yielded a single sentence for each of the 20 unigram and 20 trigram tranches.
Two additional sentences were chosen from the TIMIT corpus \citep{garofolo1993}.
Initial audio stimuli were read by a male speaker at a normal conversational pace in a soundproof environment.

All chains are initialized with a grammatical, semantically interpretable sentence, but we make no explicit effort to maintain either property over the course of serial reproduction besides the requirement for intelligibility by downstream participants.
Because later recordings may not be grammatical sentences, we refer to them as utterances, though they are sentences in a high proportion of cases (see Discussion). 

%[ ] Why are the initial sentences sampled with respect to two of the many models?
%!!! clarify grammatical, semantically well-formed
%!!! sentences, though we do not restrict that in further reproductions

% TABLE: initial stimuli examples
\begin{table*}[t]
\caption{Example transcriptions of initial stimuli, their unigram and trigram probabilities, percentile ranks, and their character length.\label{tab:inputSentences}}
\vspace{3mm}
\centering
\begin{tabular}{ l | c c c c c }
& \multicolumn{2}{ c }{Unigram} & \multicolumn{2}{ c }{Trigram} & \\
Sentence	& Log Prob. & Percentile & Log Prob. & Percentile & Characters\\
\hline
\makecell{``they found that they had \\ many of the same interests''} & -37.28  & 94 & -27.09 & 92 & 51 \\
\makecell{``the molecules that make up \\ the matter do not change''} & -38.71  & 79 & -28.51 & 85 & 51 \\
\makecell{``they went on a short hike \\ one warm autumn afternoon''} & -42.72  & 18 & -35.89 & 25 & 51 \\
\makecell{``each nonfiction book has a \\ call number on its spine''}	& -43.56  &	11 & -41.01 & 5 & 51 \\
\end{tabular}
\end{table*}

\subsubsection*{Serial Transmission Structure}

\begin{figure}[t]
\begin{center}
\includegraphics[width=4.5in]{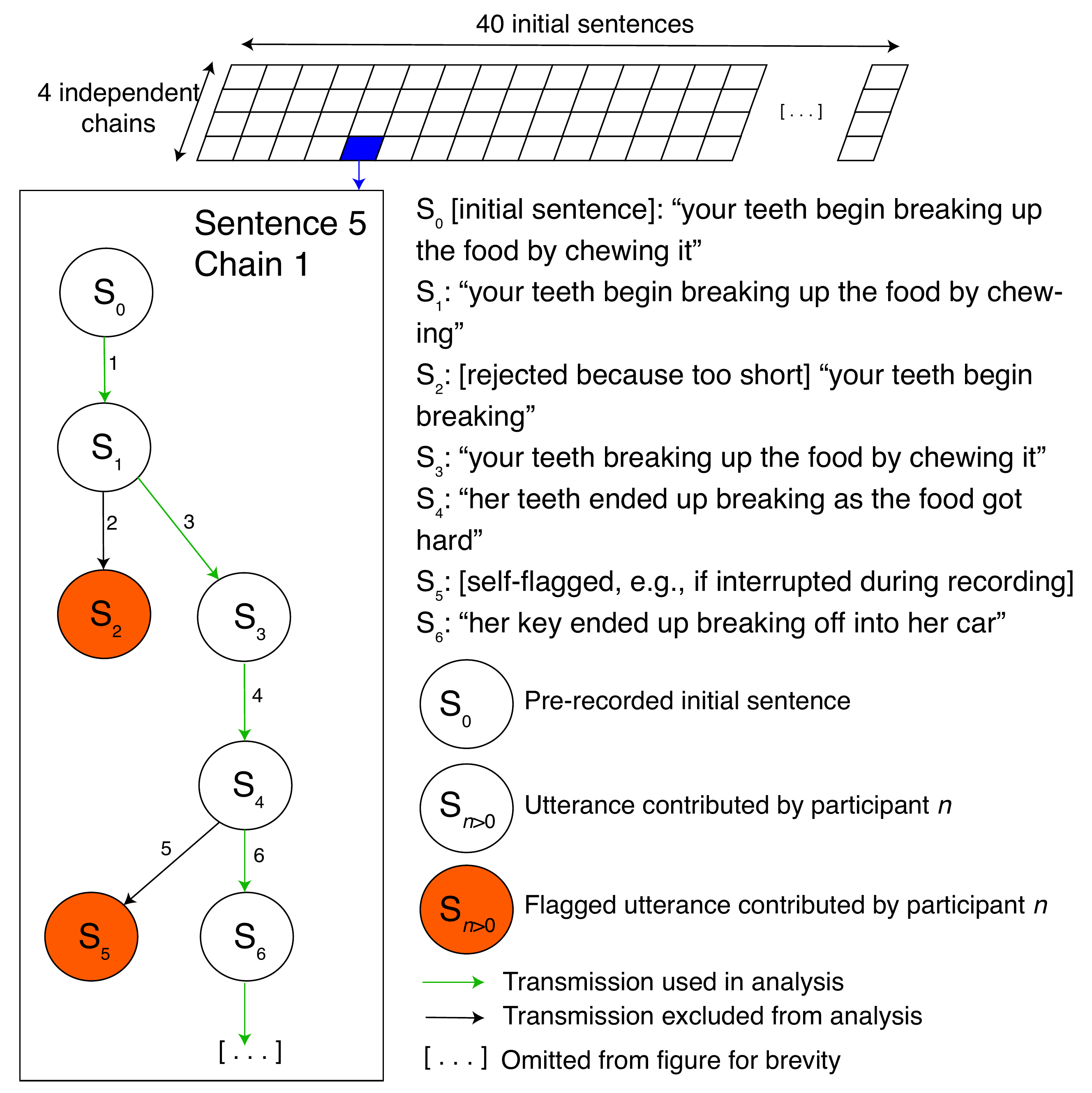}
\end{center}
\vspace{-5mm}
\caption{\label{fig:telephoneDAG} 
Schematic of the transmission structure for a single sentence in the experiment.
Participants contribute recordings according to the current state of each utterance, unlike generation-based designs. 
The web app maintains a record of which recordings are flagged by participants or automated methods as too short, bad recordings etc., and offers the last non-flagged recording  as input to a new participant.
}
\end{figure}

The web application included logic for routing newly-contributed recordings to other, later participants in the experiment.
Iterated learning and serial transmission experiments typically use a strict generation-based design where each participant contributes data in response to all relevant experiment stimuli, after which they are given to another participant.  We implemented a system that tracked the succession of inputs for each sentence separately for two reasons.
First, this increases the throughput of the experiment: under a strict generation-based design, a new participant cannot start until the previous participant has finished.
Second, it allows the experiment to balance the contributions of participants across chains in response to real time flagging of audio files (e.g., for a blank audio files).
Sequence of utterances in a chain satisfy the Markov condition because participants sequentially contribute one utterance each.
However, unlike a strict generation-based designs, two participants can provide inputs (i.e., be the immediately preceding participant) for one another on different sentences.
An example history of a sentence in the experiment, including contributions of new recordings and flagging, is presented in Figure \ref{fig:telephoneDAG}.
For more details on the transmission structure, see \textit{Technical Appendix 3: Experiment Transmission Structure}.

% FIGURE: screenshot of the experiment

\subsubsection*{Participants and Protection of Human Subjects}
A total of 266 participants were recruited using Amazon Mechanical Turk.
Participants were limited to those living in the United States with a fast internet connection, a microphone, and speakers.
Each IP address and worker identifier (temporarily stored locally in hashed form) was allowed to participate only once.
The quality of each participant's internet connection was screened at the beginning of the experiment to avoid possible issues with downloading and uploading relatively bandwidth-intensive audio.
Data collection methods, including the audio data retention and distribution policy, were reviewed and approved by the U.C. Berkeley Committee for Protection of Human Subjects.
In addition to providing informed consent, participants also completed a media release allowing their submitted audio recordings (which constitute personally-identifiable information) to be used in publicly-available corpora.
With the exception of the audio recordings, Mechanical Turk worker identifiers, and IP addresses (stored in a hashed format and discarded after the completion of the experiment), no other personally-identifiable information was collected.
Participants were not explicitly told that the recordings that they heard might come from other participants, though the media release states that their recordings \textit{could} be used as experimental stimuli.
The introduction to the experiment stated that the task was designed to gather data on how people recognize words in conditions with high levels of background noise.

%\subsubsection*{Technical Implementation}
%Additional details regarding the technical implementation are provided in the SI.
% Python Flask app on EC2; Celery + Nvidia CUDA with Tensorflow; anonymized subject data kept in Firebase
%Our code for this experimental interface is available on Github [[]]/
%[ ] Wallace / Dallinger relationship
%
%\subsubsection*{Artificial Agents}
%
%
%We conduct the same telephone experiment with artificial agents using state-of-the-art speech recognition and synthesis to examine whether the revealed inductive biases to test whether such systems are using qualitatively similar biases to people.
%These agents combine a speech recognition model (DeepSpeech, using a 5-gram language model) with a speech synthesis model (DC-TTS) such that they can contribute to chains of inputs and outputs in a similar way to human participants.
%
%\textcolor{red}{\noindent[[Details of  DeepSpeech architecture, parameter search for DeepSpeech, DC-TTS]]}

\section{Results}

The final recording chains consist of 2,999 utterances. 
An analysis of participants' flagging behavior is presented in \textit{Technical Appendix 4: Analysis of Flagged Recordings}.
In this section, we evaluate a suite of probabilistic generative language models in their ability to predict the changes made by participants in a large web-based game of ``Telephone.''
We begin by confirming that the parallel sequences of responses elicited in the telephone game demonstrate movement towards a consistent set of linguistic expectations.
We then evaluate predictive language models with two different techniques.
First, we evaluate which language models demonstrate the highest magnitude increase in probability over the course of the experiment.
Second, we test whether surprisal under each model is predictive of which words are transmitted successfully from speaker to listener.
%!!!
%Finally, we look at how chains produced by automated agents from state-of-the-art word recognition and synthesis models compare with those produced by human participants.

% Figure
%latex.default(rdf, file = file, title = "", table.env = FALSE,     booktabs = TRUE, ...)%
\begin{figure}[t]
\begin{center}
\includegraphics[width=3.5in]{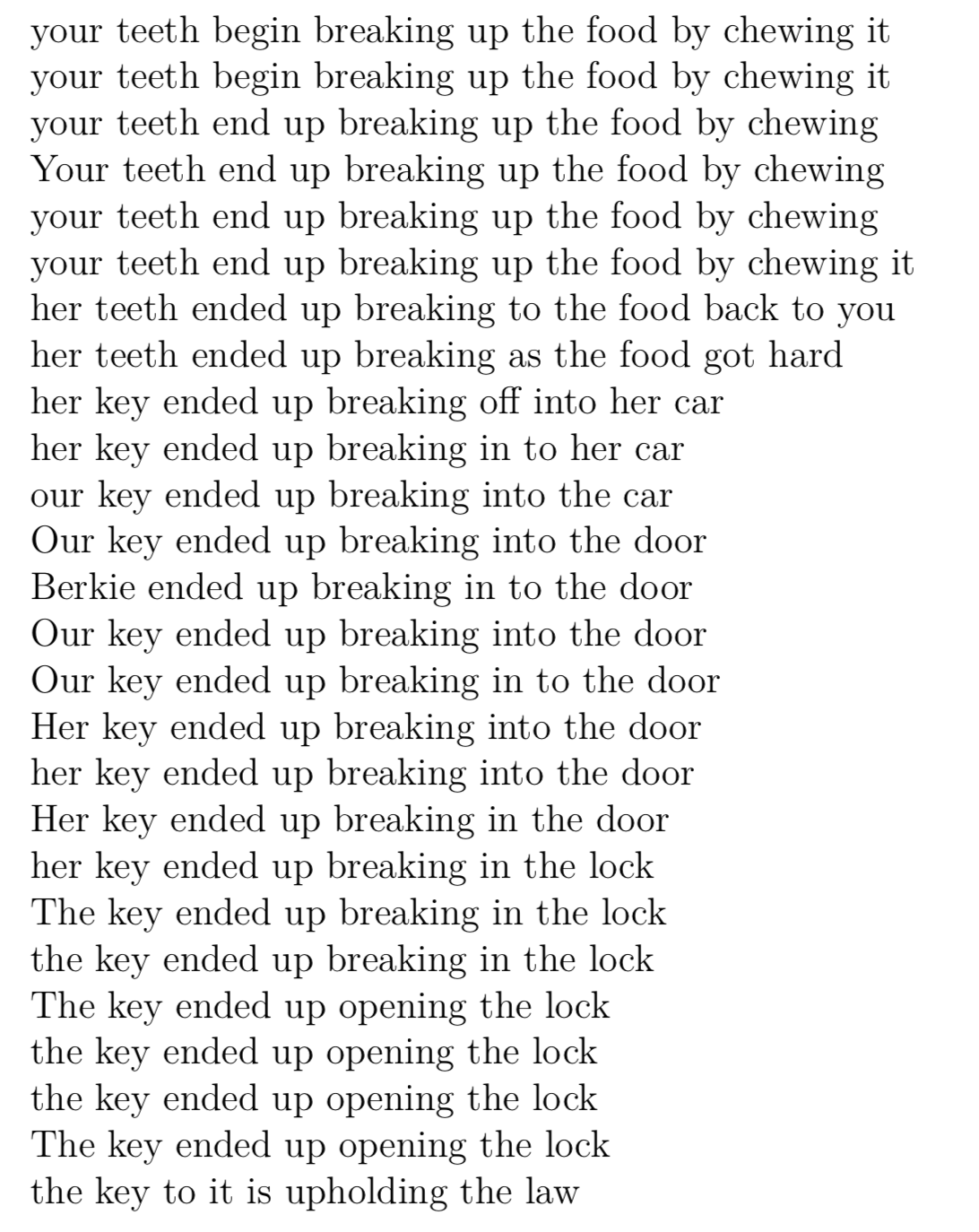}
\end{center}
\vspace{-5mm}
\caption{\label{fig:recording_chain} 
A representative recording chain yielded by the serial reproduction experiment.
The first transcription is the initial stimulus.
Each subsequent transcription is that of a participant who heard the preceding sentence presented in naturalistic background noise.
All responses are collected as audio recordings first, then participants are prompted to provide a written transcription.
}
\end{figure}

\subsection*{Evaluating Movement Towards Convergence}

We first evaluate whether the recording chains change in a way that suggests that they are headed towards convergence.
Convergence among sampling chains in Markov chain Monte Carlo is often evaluated in terms of a \textit{potential scale reduction factor}, or PSRF \citep{gelmanRubin1992}.
The PSRF measures the degree to which between-chain variance in parameter estimates reduces with respect to within-chain variance over the course of sampling.
%https://blog.stata.com/2016/05/26/gelman-rubin-convergence-diagnostic-using-multiple-chains/
In this case, we cannot directly access quantitative estimates of the parameters of the ``true'' latent model --- the linguistic expectations of human participants.
Instead we use a proxy measure:  whether the probability estimates for independent chains become more similar within a model over the course of the serial reproduction experiment.
%, and second whether the edit rates -- the probability of insertions, deletions, and substitutions -- converge 
While the probability measures could potentially become progressively more similar for other reasons, this pattern is a necessary condition that the chains are approaching the distribution of interest.
Because the model-based proxy measures above are noisy, we aggregate utterance chains into groups.
Chains are stratified into four groups on the basis of the probability quartile of the initial sentence as computed by each language model.
Initial conditions of the recording chains are known to vary significantly on this dimension, in that initial stimuli reflect a stratified sample based on  unigram and trigram probability measures.
We take the mean probability within each quartile at each generation (sequential position within the recording chain), and from that compute the inter-quartile variance.
%For the edit rate, we compute the incidence of insertion, deletion, and substitution per each quartile at each generation.

% FIGURE: convergence measurement 1: sentence probability
\begin{figure}[t]
\begin{center}
\includegraphics[width=\textwidth]{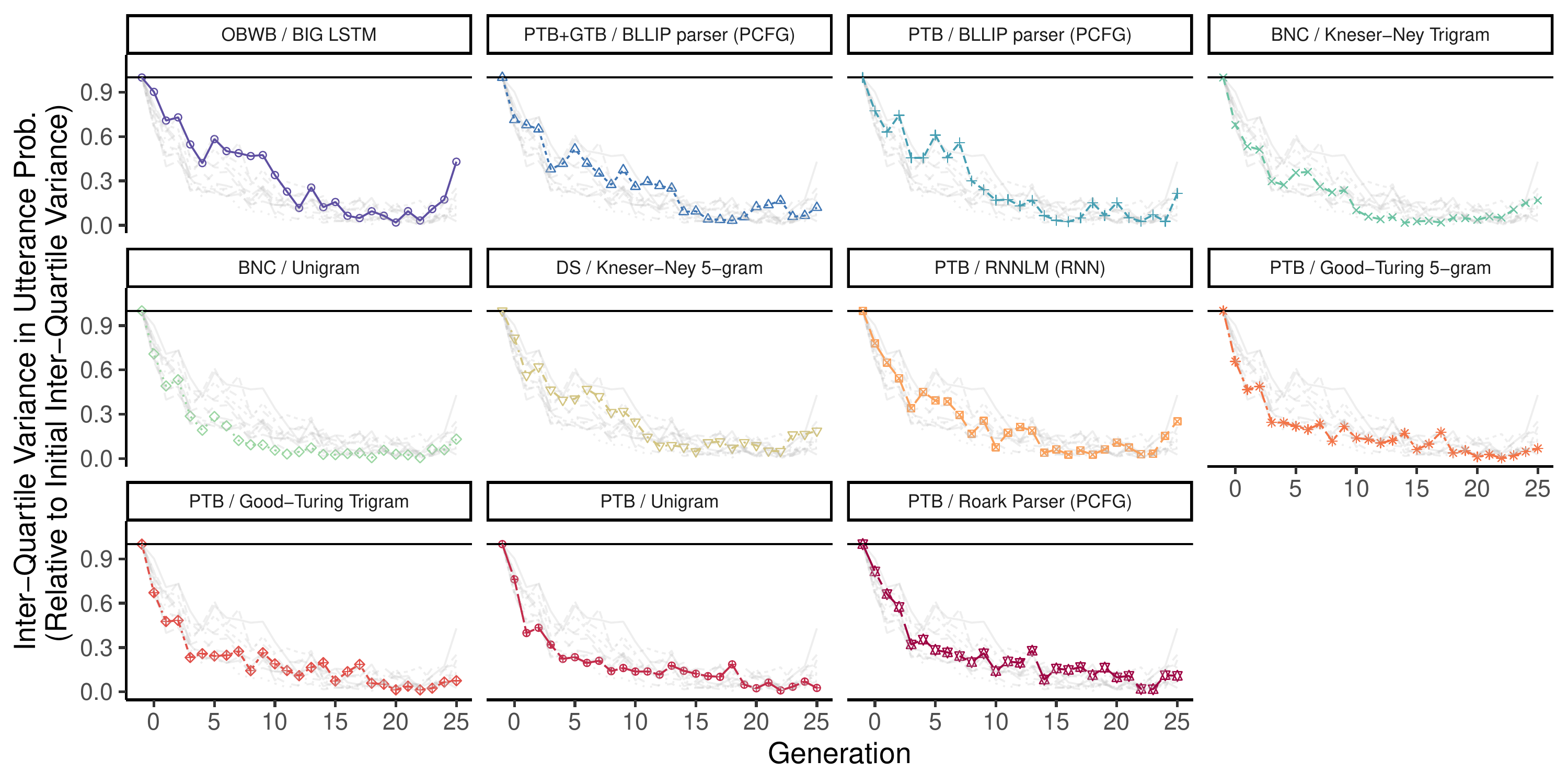}
\end{center}
\vspace{-5mm}
\caption{\label{fig:modelConvergenceAssessment} 
Variance in language probability estimates at each generation of the serial reproduction experiment relative to initial variance.
Variance estimates are computed with respect to means for four groups of chains, defined by initial sentence probability quartile.
The reduction in variance suggests that the utterances in chains starting with high probability sentences and low probability sentences becomes much more similar over the course of the experiment.
}
\end{figure}

The asymptotic decrease in inter-quartile variance seen in Figure \ref{fig:modelConvergenceAssessment} suggests that the recording chains are increasingly similar over the course of the experiment under all language models.
Figure \ref{fig:modelConvergenceAssessment} shows that variance drops to less than a quarter of the initial value for all models (with one exception for Big LM at the 25th generation), and around a tenth of the initial value for several of the models trained on large datasets ($n$-gram models trained on the BNC and the DeepSpeech datasets).
However, we caution against the stronger assertion that the chains are sampling \textit{directly} from people's expectations because utterances continue to increase in probability at the end of the experiment (next section), suggesting that the utterances need to undergo further changes to converge to human expectations.
We leave the possibility of sampling directly from the participants' revealed expectations after convergence to future research, and for now use samples generated as participants \textit{approach} the distribution of interest.

%The decrease in variance between the initial value and the final 10 generations is significant in all cases [[statistical tests]].

%Figure \ref{fig:modelConvergenceErrorRate} shows that inter-quartile variance in edit rates [. . .] over the course of serial reproduction.
% The incidence of edits provides key corroborating information in that is not dependent on a language model.

% [ ] Evaluate whether the edit probabilities converge? Use the word-level edits data that goes into the dataset -- deletions decrease but there is no motion towards convergence. Substitutions are low and constant.
% this is kind of a bummer because sentence probability was so strongly predictive of change previously  

%% FIGURE: convergence measurement 2: edit rates
%\begin{figure}
%\begin{center}
%\includegraphics[width=5.5in]{figures/modelConvergenceAssessment_edits.pdf}
%\end{center
%\vspace{-5mm}
%\caption{\label{fig:modelConvergenceErrorRate} 
%Inter-quartile variance in the rate of three types of edits -- substitutions, deletions, and insertions -- 
% at each generation of the serial reproduction experiment.
%Variance estimates are computed with respect to means for four groups of chains, defined by initial sentence probability quartile under the unigram model.
%}
%\end{figure}

\subsection*{Evaluating Language Models}

Ideally, the language models here could be evaluated  using the probability each assigns to a large set of utterances sampled from participants after strong evidence that sampling chains have fully converged to participants' expectations---both that inter-quartile variance approaches zero and that the probability of utterances is no longer increasing. 
Because utterance probabilities continue to increase, these chains are likely not directly sampling from participants' expectations by the end of the experiment.

Instead, we use the fact that serial reproduction yields sentences that are increasingly representative of people's linguistic expectations in the task, even if they have not yet converged.
As such, the pattern of changes towards the target distribution can be used to evaluate models: more representative models should have a greater magnitude increase in probability (manifested as a decrease in average per-word surprisal) over the course of the experiment.
A model reflecting the true expectations of participants should exhibit the largest possible decrease in surprisal over the course of serial transmission. 
Though we lack direct access to people's expectations, the magnitude of the increase in utterance probability for each language model---as an approximation of those human expectations-- is sufficient to rank it with respect to others.

\begin{figure}[t]
\begin{center}
\includegraphics[width=\textwidth]{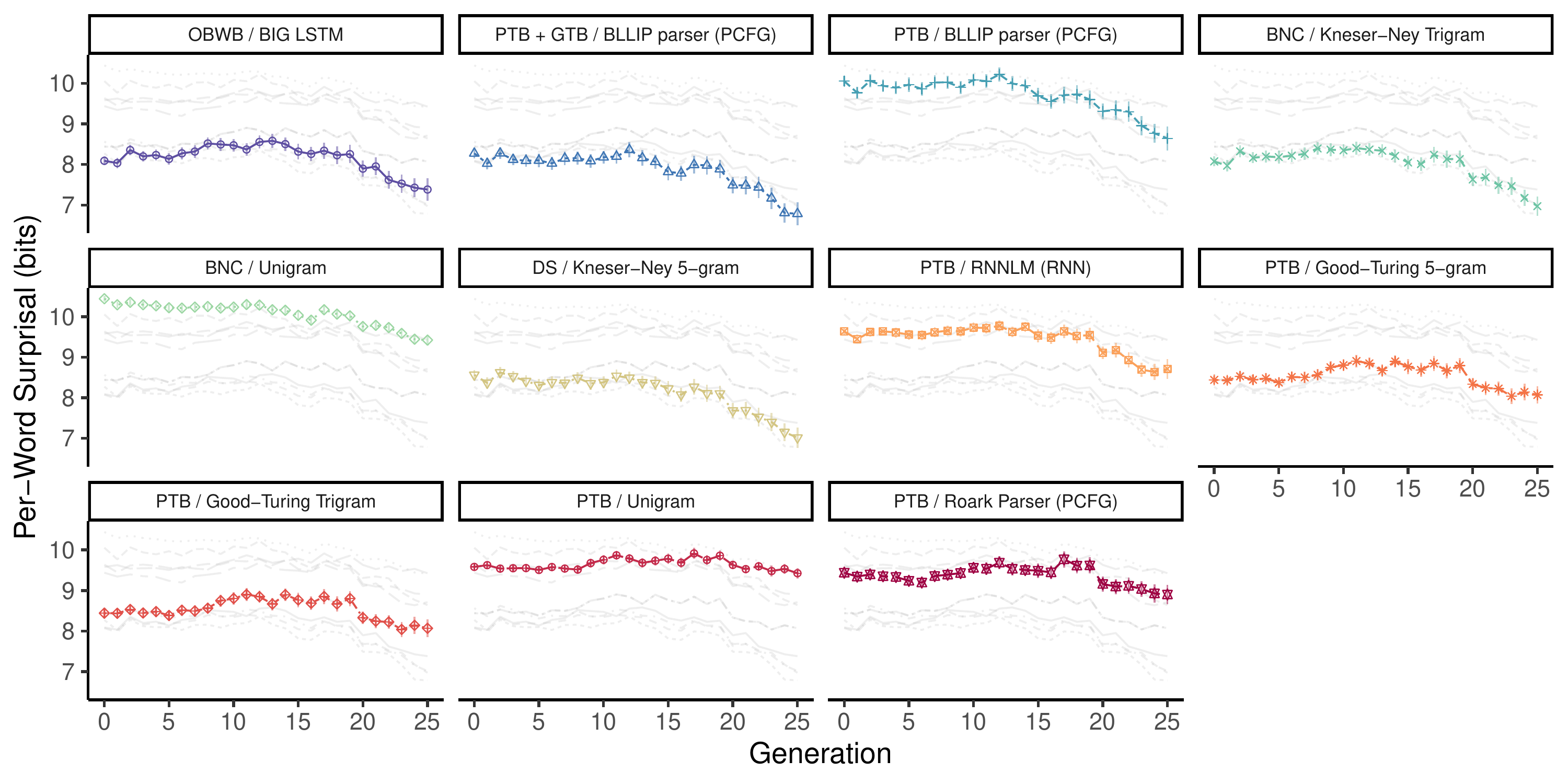}
\end{center}
\vspace{-5mm}
\caption{\label{fig:modelProbabilityComparison_raw} 
Average per-word surprisal in bits ($-\log_2(p(w|c)$) under each language model over the course of serial reproduction.
Error bars indicate standard error of the mean.
}
\end{figure}

To eliminate the effect of shorter utterances on probability, which would trivially be assigned higher probabilities under all language models, we divide each utterance's negative log probability by the number of words in that utterance.
The resulting measure, ``average per-word surprisal'' has an intuitive interpretation as the average surprisal, measurable in bits, for each utterance under each model.
The average of this measure across chains over the course of the experiment is shown in Figure \ref{fig:modelProbabilityComparison_raw}.

% FIGURE: comparison of the probability of the prior under the language models
 
Models vary in their surprisal estimates for reasons outside of the scope of the current analysis, especially the choice of smoothing scheme.
For example, consider two unigram models that withhold different amounts of probability mass for word tokens not encountered during training, e.g., .05 and .01.
If these two models were used to produce probability estimates for a set of sentences comprised of exclusively known tokens, the first model would assign a lower probability estimate compared to the second model, even though the probability estimates would be perfectly correlated.
We thus focus not on the intercept for each model but the slope of the change in surprisal, taking advantage of the fact that the change in the number of bits ($\log_{2}$ $p(w)$) corresponds to a constant multiplicative factor in probabilities (an increase of one bit, whether it is the third or the seventh, means that the words in the set of utterances are, on average, half as probable).

%This means that we measure each model's increase in probability over the course of recordings.
%This is distinguished from the common approach in model evaluation, both in Cognitive Science and NLP, which seeks the highest probability model (with various schemes to punish excessive numbers of free parameters).

% FIGURE: comparison of the probability of revealed human expectations under the language models
% normalized by convergence
%\begin{figure}[t]
%\begin{center}
%\includegraphics[width=5.5in]{figures/modelProbabilityComparison_normalized15.pdf}
%\end{center}
%\vspace{-5mm}
%\caption{\label{fig:modelProbabilityComparison15} 
%Log likelihood ratio of the recording chains under each language model, reflecting the probability of the chains at each generation relative to the probability at convergence.
%After convergence, the rate of increase in this measure (i.e., towards higher likelihood ratios) indicates stronger similarity to human linguistic expectations.  
%}
%\end{figure}

\begin{figure}[t]
\begin{center}
\includegraphics[width=\textwidth]{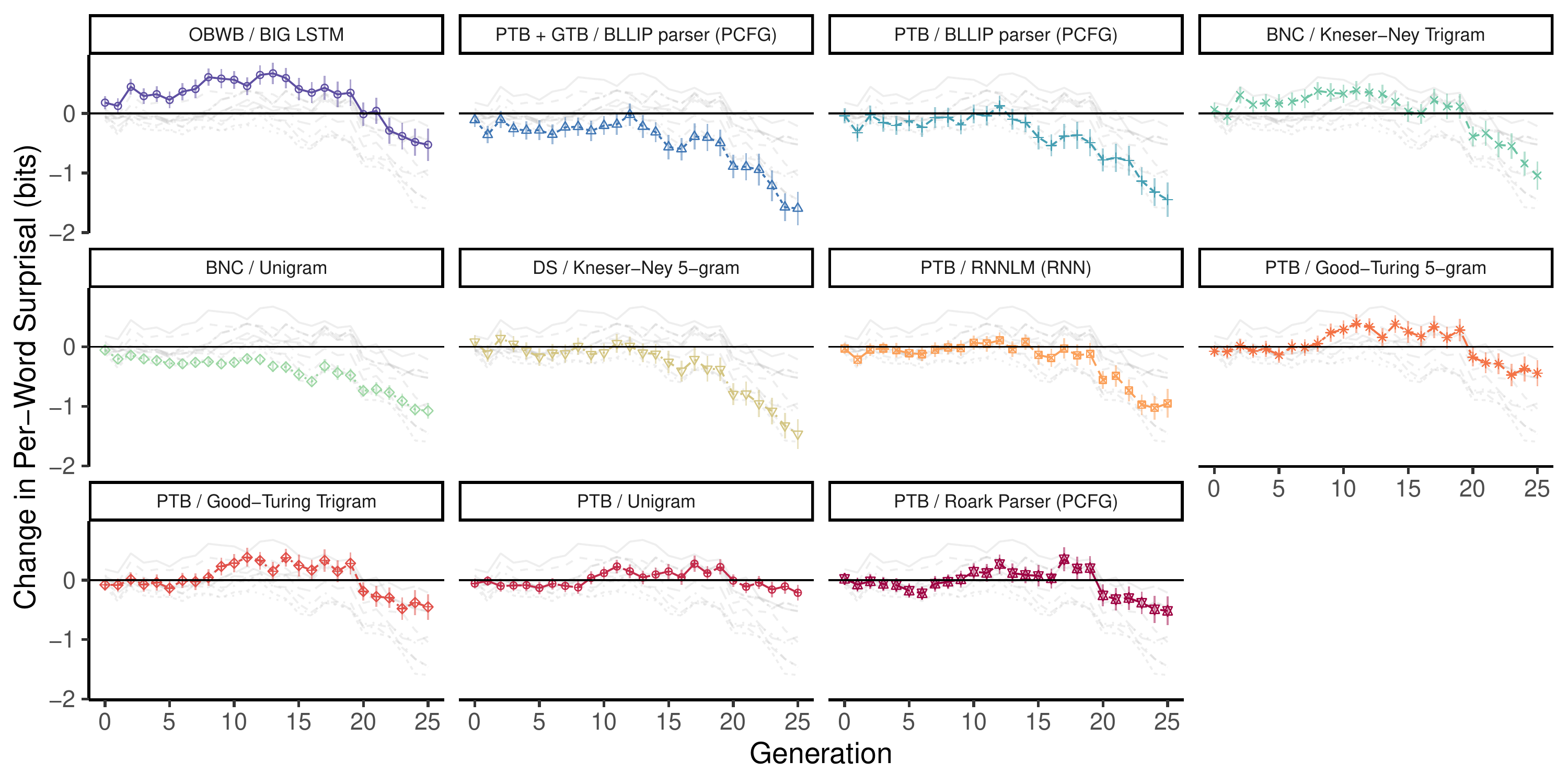}
\end{center}
\vspace{-5mm}
\caption{\label{fig:modelProbabilityComparison_normalized} 
Change in the average per-word surprisal for utterances over the course of the telephone game under 11 language models.
A decrease in average per-word surprisal of one bit means that words in the utterances yielded from serial reproduction are on average twice as probable under that model. 
}
\end{figure}

%The natural log of the likelihood ratio at each generation for each model is shown in Figure \ref{fig:modelProbabilityComparison_raw}.
While average per-word surprisal under each model for recording chains through 25 sequential transmissions is plotted in Figure \ref{fig:modelProbabilityComparison_raw}, the change in average per-word surprisal (in bits) is plotted in Figure \ref{fig:modelProbabilityComparison_normalized}.
This latter graph reveals that higher-order $n$-gram models trained on large datasets and the BLLIP PCFG parser trained on the Penn Treebank best capture the changes observed in the course of serial reproduction.
The observed pattern of increases in probability corresponds with the absolute model probabilities, in that these models are the same ones that assign the highest probability to the final generations of the sampling chains.
$n$-gram models and the Roark parser (trained on the same TreeBank dataset) show statistically significant --- though more modest --- increases in probability for later sentences.
Big LM exhibits low initial average per-word surprisal, but exhibits a relatively minor increase in probability for later sentences.

% FIGURE: comparison of the probability of revealed human expectations under the language models
% normalized by convergence

We next turn to the question of whether there are differences in performance between broad classes of models as a result of theoretically-relevant architectural distinctions, in particular whether models that represent hierarchical syntactic relationships more accurately reflect the changes made in the course of the serial reproduction experiment.

To confirm that the identified theoretical contrast (usage of abstract higher-order representations) is indeed reflected in the probability estimates produced by the models, we first measure the similarity between language models on the basis of the probability estimates they provide for all $n$ = 3,193 utterance transcriptions yielded by the experiment.
If model performance is sensitive to this distinction, correlations should be strongest within the two partitions (models that make use of hierarchical representations vs. those that do not).
Because the form of the relationship may not be linear, we use Spearman's rank correlation coefficient to evaluate the pairwise similarity.
We limit this analysis to variants of models trained on the Penn TreeBank to eliminate the effect of training dataset on model performance (see \textit{Technical Appendix 5: Similarity of Language Models} for a comparison of all models).

The results of this analysis yield two major clusters, the $n$-gram models, which do not use higher-order representations of lexical context for prediction, and the RNN-LM and the two PCFG models which do (Fig. \ref{fig:lm_similarity}).
Among the $n$-gram models, the unigram model is distinguished from the higher-order $n$-gram models; these longer-context $n$-gram models are more similar than the unigram model to the models with higher-order abstract structure.
The yielded similarities show that the predictions derived from these models reflect this key architectural distinctions of theoretical interest regarding the representation of linguistic context.
Further, these results substantiate our claim that recurrent neural network language models are better grouped with the PCFG models than the $n$-gram models given their representational capacities, in contrast to the classification used in \citet{frankBod2011} and \citet{fossumLevy2012}.

\begin{figure}[t]
\begin{center}
  \begin{subfigure}[b]{0.45\textwidth}
    \includegraphics[width=\textwidth]{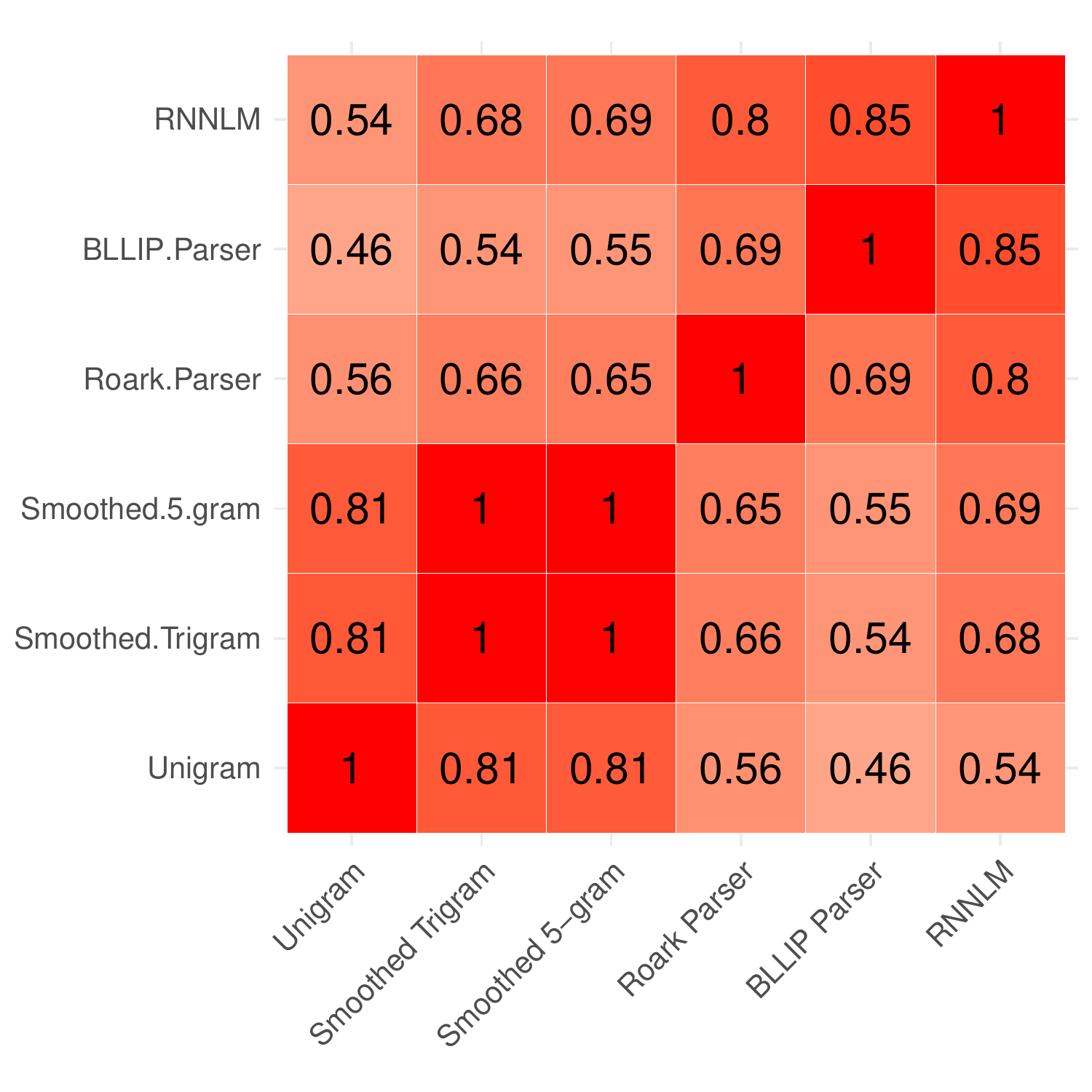}    
  \end{subfigure}
  \begin{subfigure}[b]{0.45\textwidth}
    \includegraphics[width=\textwidth]{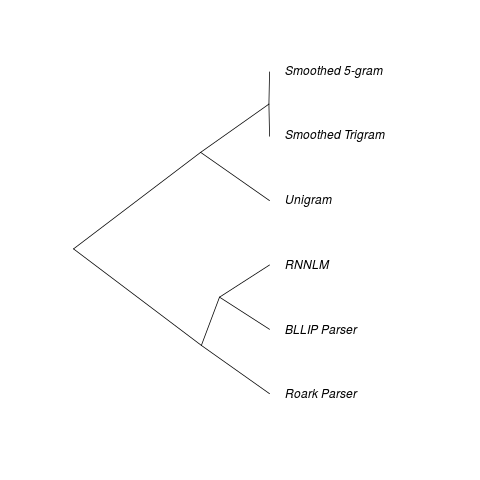}
    
  \end{subfigure}
  \caption{\textit{A.} Spearman's rank correlation for pairwise combinations of sentence probabilities across the 3,193 sentences from the experiment. \textit{B.} The resulting dendrogram, derived from the rank correlations using Ward's method, shows that the model-based probability estimates reflect the architectural difference of interest---whether models represent hierarchical syntactic relations.}
  \label{fig:lm_similarity}
 \end{center}
\end{figure}

\newcommand{\structureTestsIntercept}{($\beta$ = 2.3362, \textit{t} value = 85.81, $p$ $<$ .0001)}
\newcommand{\structureTestsgeneration}{($\beta$ = 0.001, \textit{t} value = 0.38, $p$ = 0.7)}
\newcommand{\structureTestsabstractstructureused}{($\beta$ = 0.2382, \textit{t} value = 25.8, $p$ $<$ .0001)}
\newcommand{\structureTestsdatasetPTB}{($\beta$ = 0.2923, \textit{t} value = 33.38, $p$ $<$ .0001)}
\newcommand{\structureTestsgenerationxabstractstructureused}{($\beta$ = -0.0044, \textit{t} value = -6, $p$ $<$ .0001)}
\newcommand{\structureTestsgenerationxdatasetPTB}{($\beta$ = 0.0042, \textit{t} value = 6.03, $p$ $<$ .0001)}
\newcommand{\structureTestsInterceptRecordingChain}{(Std. Dev. = 0.35}
\newcommand{\structureTestsgenerationRecordingChain}{(Std. Dev. = 0.03}

The central architectural question we address is whether models that represent higher-order regularities, such as phrase structure or abstract semantic representations of the preceding context, better reflect the revealed linguistic expectations of participants than models that track only specific preceding words.
The $n$-gram models (where $n$ $>$ 1) use purely lexicalized representations of preceding context, and do not posit any sort of higher-order abstractions such as verb phrases or noun phrases.
By contrast, the PCFGs explicitly represent phrase structure, and recurrent neural network language models are sensitive to higher-order syntactic and semantic regularities insofar as these are captured in the lower-dimensional embedding of preceding context.

We construct a mixed effects linear regression model to predict average per-word surprisal with the following predictors: dataset (Penn TreeBank vs. large representative dataset; treatment coded with the latter as the reference level), abstract representation of context (used vs. not used; treatment coded with the latter as the reference level), and the interaction of both preceding terms with generation.
Unique recording chains (independent sequences of utterances from serial reproduction) are treated as random intercepts, and we account for a \texttt{recording chain} $\times$ \texttt{generation} random slope.  
The regression model is fit with 11 model estimates of average per-word surprisal for each of 2,864 utterances produced between the 1st and 26th participant (thereby omitting the initial sentences).

We first test whether accounting for each language model's type of representation (abstract hierarchical vs. lexical) improves the overall fit of the mixed-effects model by comparing the above full model specification with a nested model lacking type of context representation as a predictor.
This reveals that the model that includes the type of context representation as well as its interaction with generation exhibits significantly better fit ($\chi^{2}$ = 1401.1, p $<$.0001)
For this full model (Tab. \ref{tab:structure_lm}), there is a statistically significant negative coefficient for the interaction for \texttt{generation} * \texttt{abstract~structure:~used} \structureTestsgenerationxabstractstructureused.
This greater magnitude decrease in surprisal estimates suggest that the changes made by people are better reflected by the language models that use abstract representations of preceding context.
As with the previous analysis, we find that models that are trained on the Penn TreeBank are less representative of people's linguistic expectations than models trained on larger datasets \structureTestsgenerationxdatasetPTB.

%latex.default(rdf, file = file, title = "", table.env = FALSE,     booktabs = TRUE, ...)%
\begin{table}[t]
\caption{A mixed-effects linear regression examining average per-word surprisal estimates (negative log probability under a model) across 9 language models as a function of 1) whether the model uses an abstract representation of the preceding context to inform expectations and 2) the dataset used to fit the model.\label{tab:structure_lm}} 
\begin{center}
\begin{tabular}{lllll}
\toprule
Fixed Effects&&&&\tabularnewline
\multicolumn{1}{l}{}&\multicolumn{1}{c}{Coef $\beta$}&\multicolumn{1}{c}{SE($\beta$)}&\multicolumn{1}{c}{\textit{t} value}&\multicolumn{1}{c}{$p$}\tabularnewline
\midrule
(Intercept)&2.3362&0.0272&85.81&\textbf{\textless .0001}\tabularnewline
generation&0.001&0.0025&0.38&0.7\tabularnewline
abstract structure: used&0.2382&0.0092&25.8&\textbf{\textless .0001}\tabularnewline
dataset: PTB&0.2923&0.0088&33.38&\textbf{\textless .0001}\tabularnewline
generation x abstract structure: used&-0.0044&7e-04&-6&\textbf{\textless .0001}\tabularnewline
generation x dataset: PTB&0.0042&7e-04&6.03&\textbf{\textless .0001}\tabularnewline
\hline
&&&&\tabularnewline
Random Effects&&&&\tabularnewline
&Std. Dev&&&\tabularnewline
(Intercept) $|$ Recording Chain&0.35&&&\tabularnewline
generation $|$ Recording Chain&0.03&&&\tabularnewline
\bottomrule
\multicolumn{5}{l}{\footnotesize{Note: Significance of fixed-effects is computed following Satterthwaite, 1946.}}
\end{tabular}\end{center}
\end{table}

\subsubsection*{Predicting Word-Level Errors}

A second way to compare the fit of probabilistic generative language models to human behavior is to evaluate how well they predict which words are misheard: altered or deleted in transmission.\footnote{While we also collect data regarding insertions in the course of serial reproduction, it is harder to identify the relevant properties of the preceding recording that prompt the insertion of material.}
We specifically test whether models that represent abstract hierarchical structure provide predictive power, in the form of their conditional probability estimates, beyond lexical models regarding which words are altered or deleted.
Instances of deletions and substitutions are identified using dynamic programming, using the edit operations corresponding to the Levenshtein distance, or minimum edit distance between utterances.
As is commonly done to estimate Word Error Rate (WER) in natural language processing \citep{popovicNey2007}, this computation is applied to word sequences rather than character strings (Tab. \ref{tab:MDIM}).
We present a separate analysis of participants' edit rates in relation to their flagging behavior in 
 \textit{Technical Appendix 4: Analysis of Flagged Recordings}.

We then construct a mixed-effects logistic regression model including model-based estimates of surprisal, as well as the control variables of position in sentence, age of acquisition ratings \citep{kupermanEtAl2012}, concreteness \citep{brysbaerEtAlt2014}, number of phonemes, number of syllables \citep{brysbaertNew2009}, and phonological neighborhood density \citep{yarkoniEtAl2008} as fixed effects to predict whether the word changes (1 = change, 0 = no change).
We restrict model based estimates to those from larger datasets (i.e., omitting models trained on the Penn TreeBank if another model with the same architecture but trained on a larger dataset is available).
To deal with the strong correlations between language model predictions, we use a residualization scheme to isolate their respective contributions.
Unigram surprisal (negative log probability) is used directly as a predictor.
We then predict trigram probability from unigram probability, and use the residuals as a predictor representative of the contribution of a language model which considers the two words preceding the words in question as the preceding context.
For each of the remaining models with word-level surprisal estimates (5-gram on the DeepSpeech dataset, BigLM on the One Billion Word Benchmark, the Roark Parser trained on the Penn TreeBank) we take the residuals after predicting its surprisal estimates from both unigram and trigram models.

The identities of the listener and the identity of the speaker are treated as random intercepts to account for variability in comprehension performance and speaker intelligibility.
The model is fit with 27,290 instances where a participant heard a word and either 1) reproduced it faithfully in their own recording (22,482 cases) 2) produced an identifiable substitute (substitution) or 3) did not produce an identifiable substitution (deletion).
Cases 2) and 3) were collapsed into a single category of transmission failure, coded as a 0.
We fit the full model and conduct no pruning of predictors.

% TABLE: MDIM example
\begin{table}[t]
\centering
\caption{Example edit string from input sentence to output sentence. M, D, I, and S indicate Match, Deletion, Insertion, and Substitution respectively.} 
\vspace{3mm}
\setlength{\tabcolsep}{.16667em}
\begin{tabular}{c|c|c|c|c|c|c|c|c|c|c|c}
M & M & M & M & D & I & I & M & D & M & S & M \\
\hline
\cellcolor[gray]{0.8}you & may &\cellcolor[gray]{0.8}not & notice & \cellcolor[gray]{0.8}yourself & &\cellcolor[gray]{0.8}  &grow& \cellcolor[gray]{0.8}from & day & to & \cellcolor[gray]{0.8}day \\
\cellcolor[gray]{0.8}you & may &\cellcolor[gray]{0.8}not & notice & \cellcolor[gray]{0.8}& as & \cellcolor[gray]{0.8}you & grow &\cellcolor[gray]{0.8} & day & by & \cellcolor[gray]{0.8}day
\end{tabular}
\label{tab:MDIM}
\end{table}

\newcommand{\wordLevelIntercept}{($\beta$ = -2.6554, \textit{z} value = -28.45, $p$$<$ .0001)}
\newcommand{\wordLevelBNCunigramsurprisal}{($\beta$ = 0.3827, \textit{z} value = 17.25, $p$$<$ .0001)}
\newcommand{\wordLevelResidualizedBNCtrigramsurprisal}{($\beta$ = 0.3729, \textit{z} value = 19.37, $p$$<$ .0001)}
\newcommand{\wordLevelResidualizedRoarkPCFGsyntacticsurprisal}{($\beta$ = 0.1159, \textit{z} value = 4.36, $p$$<$ .0001)}
\newcommand{\wordLevelResidualizedBigLMsurprisal}{($\beta$ = 0.1959, \textit{z} value = 10.82, $p$$<$ .0001)}
\newcommand{\wordLevelResidualizedDSgramsurprisal}{($\beta$ = 0.1593, \textit{z} value = 6.07, $p$$<$ .0001)}
\newcommand{\wordLevelPositioninsentence}{($\beta$ = 0.0989, \textit{z} value = 15.57, $p$$<$ .0001)}
\newcommand{\wordLevelAgeofacquisition}{($\beta$ = 0.0544, \textit{z} value = 3.96, $p$$<$ .0001)}
\newcommand{\wordLevelNumberofphonemes}{($\beta$ = -0.0609, \textit{z} value = -2.64, $p$$<$ .01)}
\newcommand{\wordLevelNumberofsyllables}{($\beta$ = -0.249, \textit{z} value = -4.91, $p$$<$ .0001)}
\newcommand{\wordLevelConcreteness}{($\beta$ = -0.0856, \textit{z} value = -4.46, $p$$<$ .0001)}
\newcommand{\wordLevelPhonologicalNeighborhoodDensity}{($\beta$ = -0.0948, \textit{z} value = -2.22, $p$$<$ .05)}
\newcommand{\wordLevelListenerID}{(Std. Dev. = 0.62}
\newcommand{\wordLevelSpeakerID}{(Std. Dev. = 0.3}

We report the contributions of lexical properties and the word's position of the sentence before reporting on the contribution of model-based estimates of predictability.
Words are more likely to change if they are appear later in the utterance \wordLevelPositioninsentence, or if that word is acquired later in development \wordLevelAgeofacquisition.
By contrast, words with more syllables or more phonemes are \textit{less} likely to change \wordLevelNumberofsyllables, as are words rated as highly concrete \wordLevelConcreteness.
In line with recent regression models investigating recognition accuracy in isolated word \citep{gahlStrand2016}, this analysis suggests that words in sparse phonological neighborhoods --- words with high average phonological Levenshtein distance to the 20 most similar wordforms (or PLD-20, per \citep{yarkoniEtAl2008}) are less likely to change \wordLevelPhonologicalNeighborhoodDensity.

\begin{figure} 
\centering
     \begin{subfigure}{0.2\textwidth}
     \end{subfigure}
	 \begin{subfigure}{0.5\textwidth}     
\begin{flushleft}
\begin{singlespacing}
{\footnotesize
\textcolor[HTML]{966800}{your} \textcolor[HTML]{C63800}{teeth} \textcolor[HTML]{847A00}{begin} \textcolor[HTML]{6B9300}{breaking} \textcolor[HTML]{47B700}{up} \textcolor[HTML]{728C00}{the} \textcolor[HTML]{8E7000}{food} \textcolor[HTML]{EA1400}{by} \textcolor[HTML]{EF0F00}{chewing} \textcolor[HTML]{F70700}{it}
 \\ 
\textcolor[HTML]{966800}{your} \textcolor[HTML]{C63800}{teeth} \textcolor[HTML]{847A00}{begin} \textcolor[HTML]{6B9300}{breaking} \textcolor[HTML]{47B700}{up} \textcolor[HTML]{728C00}{the} \textcolor[HTML]{8E7000}{food} \textcolor[HTML]{EA1400}{by} \textcolor[HTML]{EF0F00}{chewing} \textcolor[HTML]{F70700}{it}
 \\ 
\textcolor[HTML]{966800}{your} \textcolor[HTML]{C63800}{teeth} \textcolor[HTML]{D82600}{end} \textcolor[HTML]{16E800}{up} \textcolor[HTML]{7A8400}{breaking} \textcolor[HTML]{669900}{up} \textcolor[HTML]{728C00}{the} \textcolor[HTML]{9B6300}{food} \textcolor[HTML]{ED1100}{by} \textcolor[HTML]{F40A00}{chewing}
 \\ 
\textcolor[HTML]{966800}{your} \textcolor[HTML]{C63800}{teeth} \textcolor[HTML]{D82600}{end} \textcolor[HTML]{16E800}{up} \textcolor[HTML]{7A8400}{breaking} \textcolor[HTML]{669900}{up} \textcolor[HTML]{728C00}{the} \textcolor[HTML]{9B6300}{food} \textcolor[HTML]{ED1100}{by} \textcolor[HTML]{F40A00}{chewing}
 \\ 
\textcolor[HTML]{966800}{your} \textcolor[HTML]{C63800}{teeth} \textcolor[HTML]{D82600}{end} \textcolor[HTML]{16E800}{up} \textcolor[HTML]{7A8400}{breaking} \textcolor[HTML]{669900}{up} \textcolor[HTML]{728C00}{the} \textcolor[HTML]{9B6300}{food} \textcolor[HTML]{ED1100}{by} \textcolor[HTML]{F40A00}{chewing}
 \\ 
\textcolor[HTML]{966800}{your} \textcolor[HTML]{C63800}{teeth} \textcolor[HTML]{D82600}{end} \textcolor[HTML]{16E800}{up} \textcolor[HTML]{7A8400}{breaking} \textcolor[HTML]{669900}{up} \textcolor[HTML]{728C00}{the} \textcolor[HTML]{9B6300}{food} \textcolor[HTML]{ED1100}{by} \textcolor[HTML]{F40A00}{chewing} \textcolor[HTML]{F90500}{it}
 \\ 
\textcolor[HTML]{827C00}{her} \textcolor[HTML]{D62800}{teeth} \textcolor[HTML]{C93500}{ended} \textcolor[HTML]{23DB00}{up} \textcolor[HTML]{D82600}{breaking} \textcolor[HTML]{DB2300}{to} \textcolor[HTML]{8E7000}{the} \textcolor[HTML]{D12D00}{food} \textcolor[HTML]{EF0F00}{back} \textcolor[HTML]{7F7F00}{to} \textcolor[HTML]{ED1100}{you}
 \\ 
\textcolor[HTML]{827C00}{her} \textcolor[HTML]{D32B00}{teeth} \textcolor[HTML]{C93500}{ended} \textcolor[HTML]{23DB00}{up} \textcolor[HTML]{D82600}{breaking} \textcolor[HTML]{DD2100}{as} \textcolor[HTML]{7C8200}{the} \textcolor[HTML]{CE3000}{food} \textcolor[HTML]{E81600}{got} \textcolor[HTML]{E51900}{hard}
 \\ 
\textcolor[HTML]{827C00}{her} \textcolor[HTML]{EA1400}{key} \textcolor[HTML]{D32B00}{ended} \textcolor[HTML]{0AF400}{up} \textcolor[HTML]{DB2300}{breaking} \textcolor[HTML]{AA5400}{off} \textcolor[HTML]{8E7000}{into} \textcolor[HTML]{CC3200}{her} \textcolor[HTML]{A05E00}{car}
 \\ 
\textcolor[HTML]{827C00}{her} \textcolor[HTML]{ED1100}{key} \textcolor[HTML]{D32B00}{ended} \textcolor[HTML]{0AF400}{up} \textcolor[HTML]{DB2300}{breaking} \textcolor[HTML]{758900}{in} \textcolor[HTML]{ED1100}{to} \textcolor[HTML]{A05E00}{her} \textcolor[HTML]{916D00}{car}
 \\ 
\textcolor[HTML]{7F7F00}{our} \textcolor[HTML]{C63800}{key} \textcolor[HTML]{E51900}{ended} \textcolor[HTML]{1EE000}{up} \textcolor[HTML]{D82600}{breaking} \textcolor[HTML]{639B00}{into} \textcolor[HTML]{59A500}{the} \textcolor[HTML]{AD5100}{car}
 \\ 
\textcolor[HTML]{7F7F00}{our} \textcolor[HTML]{C63800}{key} \textcolor[HTML]{E51900}{ended} \textcolor[HTML]{1CE200}{up} \textcolor[HTML]{D82600}{breaking} \textcolor[HTML]{639B00}{into} \textcolor[HTML]{5BA300}{the} \textcolor[HTML]{E01E00}{door}
 \\ 
\textcolor[HTML]{C93500}{berkie} \textcolor[HTML]{996600}{ended} \textcolor[HTML]{07F700}{up} \textcolor[HTML]{D62800}{breaking} \textcolor[HTML]{7A8400}{in} \textcolor[HTML]{DD2100}{to} \textcolor[HTML]{5EA000}{the} \textcolor[HTML]{B24C00}{door}
 \\ 
\textcolor[HTML]{7F7F00}{our} \textcolor[HTML]{C63800}{key} \textcolor[HTML]{E51900}{ended} \textcolor[HTML]{1CE200}{up} \textcolor[HTML]{D82600}{breaking} \textcolor[HTML]{639B00}{into} \textcolor[HTML]{5BA300}{the} \textcolor[HTML]{E01E00}{door}
 \\ 
\textcolor[HTML]{7F7F00}{our} \textcolor[HTML]{C63800}{key} \textcolor[HTML]{E51900}{ended} \textcolor[HTML]{1EE000}{up} \textcolor[HTML]{D82600}{breaking} \textcolor[HTML]{8C7200}{in} \textcolor[HTML]{E51900}{to} \textcolor[HTML]{669900}{the} \textcolor[HTML]{BF3F00}{door}
 \\ 
\textcolor[HTML]{827C00}{her} \textcolor[HTML]{ED1100}{key} \textcolor[HTML]{D32B00}{ended} \textcolor[HTML]{0AF400}{up} \textcolor[HTML]{DB2300}{breaking} \textcolor[HTML]{51AD00}{into} \textcolor[HTML]{5BA300}{the} \textcolor[HTML]{D32B00}{door}
 \\ 
\textcolor[HTML]{827C00}{her} \textcolor[HTML]{ED1100}{key} \textcolor[HTML]{D32B00}{ended} \textcolor[HTML]{0AF400}{up} \textcolor[HTML]{DB2300}{breaking} \textcolor[HTML]{51AD00}{into} \textcolor[HTML]{5BA300}{the} \textcolor[HTML]{D32B00}{door}
 \\ 
\textcolor[HTML]{827C00}{her} \textcolor[HTML]{ED1100}{key} \textcolor[HTML]{D32B00}{ended} \textcolor[HTML]{0AF400}{up} \textcolor[HTML]{DB2300}{breaking} \textcolor[HTML]{758900}{in} \textcolor[HTML]{56A800}{the} \textcolor[HTML]{C63800}{door}
 \\ 
\textcolor[HTML]{827C00}{her} \textcolor[HTML]{ED1100}{key} \textcolor[HTML]{D32B00}{ended} \textcolor[HTML]{0AF400}{up} \textcolor[HTML]{DB2300}{breaking} \textcolor[HTML]{758900}{in} \textcolor[HTML]{56A800}{the} \textcolor[HTML]{EA1400}{lock}
 \\ 
\textcolor[HTML]{0CF200}{the} \textcolor[HTML]{A85600}{key} \textcolor[HTML]{EA1400}{ended} \textcolor[HTML]{35C900}{up} \textcolor[HTML]{DD2100}{breaking} \textcolor[HTML]{897500}{in} \textcolor[HTML]{54AA00}{the} \textcolor[HTML]{ED1100}{lock}
 \\ 
\textcolor[HTML]{0CF200}{the} \textcolor[HTML]{A85600}{key} \textcolor[HTML]{EA1400}{ended} \textcolor[HTML]{35C900}{up} \textcolor[HTML]{DD2100}{breaking} \textcolor[HTML]{897500}{in} \textcolor[HTML]{54AA00}{the} \textcolor[HTML]{ED1100}{lock}
 \\ 
\textcolor[HTML]{0CF200}{the} \textcolor[HTML]{A85600}{key} \textcolor[HTML]{EA1400}{ended} \textcolor[HTML]{35C900}{up} \textcolor[HTML]{DD2100}{opening} \textcolor[HTML]{A85600}{the} \textcolor[HTML]{F40A00}{lock}
 \\ 
\textcolor[HTML]{0CF200}{the} \textcolor[HTML]{A85600}{key} \textcolor[HTML]{EA1400}{ended} \textcolor[HTML]{35C900}{up} \textcolor[HTML]{DD2100}{opening} \textcolor[HTML]{A85600}{the} \textcolor[HTML]{F40A00}{lock}
 \\ 
\textcolor[HTML]{0CF200}{the} \textcolor[HTML]{A85600}{key} \textcolor[HTML]{EA1400}{ended} \textcolor[HTML]{35C900}{up} \textcolor[HTML]{DD2100}{opening} \textcolor[HTML]{A85600}{the} \textcolor[HTML]{F40A00}{lock}
 \\ 
\textcolor[HTML]{0CF200}{the} \textcolor[HTML]{A85600}{key} \textcolor[HTML]{EA1400}{ended} \textcolor[HTML]{35C900}{up} \textcolor[HTML]{DD2100}{opening} \textcolor[HTML]{A85600}{the} \textcolor[HTML]{F40A00}{lock}
 \\ 
\textcolor[HTML]{11ED00}{the} \textcolor[HTML]{A85600}{key} \textcolor[HTML]{1EE000}{to} \textcolor[HTML]{966800}{it} \textcolor[HTML]{2DD100}{is} \textcolor[HTML]{EA1400}{upholding} \textcolor[HTML]{38C600}{the} \textcolor[HTML]{A55900}{law}}
\end{singlespacing}
\end{flushleft}
\includegraphics[width=0in]{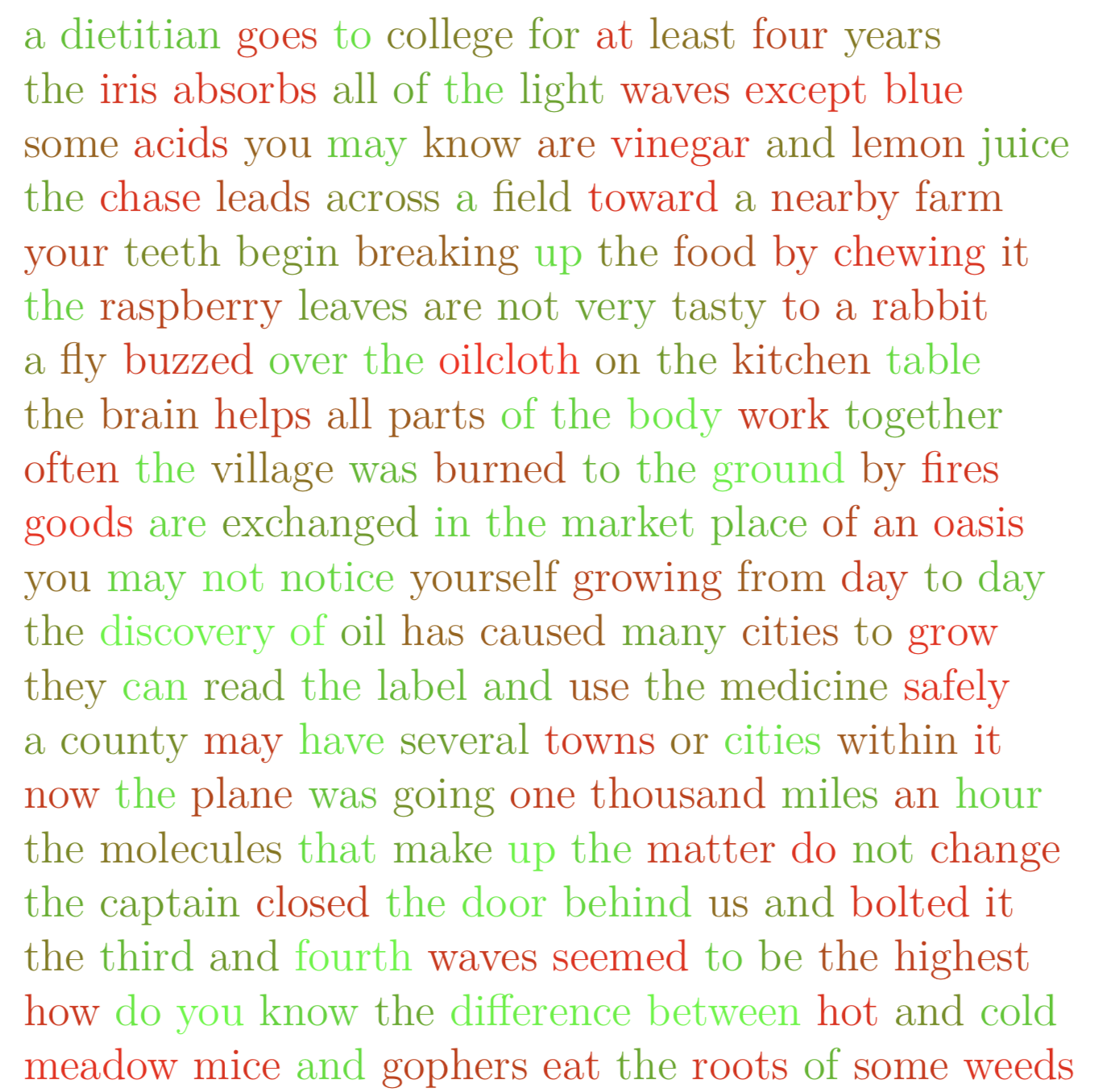}
\end{subfigure}
\begin{subfigure}{0.2\textwidth}
\end{subfigure}
\vspace{-5mm}
\caption{\label{fig:changePredicted} 
Probability that each word from an example recording chain is changed in transmission as predicted by a mixed effects logistic regression model.
Red indicates that a word is likely to be deleted or replaced with a substitution, while green suggests that a listener is likely to reproduce the word successfully. 
Colors are mapped from high to low for each sentence.
}
\end{figure}

The mixed-effects logistic regression reveals that words with higher unigram surprisal are more likely to change \wordLevelBNCunigramsurprisal, as are words with higher trigram surprisal with unigram surprisal partialed out \wordLevelResidualizedBNCtrigramsurprisal. 
For all models with abstract structure (with trigram surprisal partialed out), higher residualized surprisal estimates are predictive of a transmission failure.
This corroborates the effect of word frequency on recognition seen in \citet{gahlStrand2016}, and is consistent with the independent contributions of word frequency and word predictability seen in automatic speech recognition systems  \citep{goldwaterEtAl2010}.
An example transmission chain with each word colorized by the probability of successful recovery by the listener under the full model  are shown in Fig. \ref{fig:changePredicted}. 
This figure shows higher rates of transmission failures for low-frequency ``content'' words vs. generally high-frequency ``function'' words, though function words in certain constructions have higher probabilities of transmission failure (e.g. ``\textit{in to} the door'').

%latex.default(rdf, file = file, title = "", table.env = FALSE,     booktabs = TRUE, ...)%
\begin{table}[t]
\caption{Mixed-effects logistic regression predicting whether a word will be changed in transmission on the basis of its surprisal under various language models as well as other word properties.\label{tab:wordlevel_lm}} 
\begin{center}
\begin{tabular}{lllll}
\toprule
Fixed Effects&&&&\tabularnewline
\multicolumn{1}{l}{}&\multicolumn{1}{c}{Coef $\beta$}&\multicolumn{1}{c}{SE($\beta$)}&\multicolumn{1}{c}{\textit{z} value}&\multicolumn{1}{c}{$Pr(>|\textit{z}|)$}\tabularnewline
\midrule
(Intercept)&-2.6554&0.0933&-28.45&\textbf{\textless .0001}\tabularnewline
BNC unigram surprisal&0.3827&0.0222&17.25&\textbf{\textless .0001}\tabularnewline
Residualized BNC trigram surprisal&0.3729&0.0193&19.37&\textbf{\textless .0001}\tabularnewline
Residualized Roark PCFG syntactic surprisal&0.1159&0.0266&4.36&\textbf{\textless .0001}\tabularnewline
Residualized Big LM surprisal&0.1959&0.0181&10.82&\textbf{\textless .0001}\tabularnewline
Residualized DS 5-gram surprisal&0.1593&0.0262&6.07&\textbf{\textless .0001}\tabularnewline
Position in sentence&0.0989&0.0064&15.57&\textbf{\textless .0001}\tabularnewline
Age of acquisition&0.0544&0.0137&3.96&\textbf{\textless .0001}\tabularnewline
Number of phonemes&-0.0609&0.0231&-2.64&\textbf{\textless .01}\tabularnewline
Number of syllables&-0.249&0.0507&-4.91&\textbf{\textless .0001}\tabularnewline
Concreteness&-0.0856&0.0192&-4.46&\textbf{\textless .0001}\tabularnewline
Phonological Neighborhood Density (PLD20)&-0.0948&0.0427&-2.22&\textbf{\textless .05}\tabularnewline
\hline
&&&&\tabularnewline
Random Effects&&&&\tabularnewline
&Std. Dev.&&&\tabularnewline
\hline
Listener ID&0.62&&&\tabularnewline
Speaker ID&0.3&&&\tabularnewline
\bottomrule
\multicolumn{5}{l}{\footnotesize{Note: Significance of fixed-effects is computed following Satterthwaite, 1946.}}
\end{tabular}\end{center}
\end{table}

Given that the fit model can be used to estimate the probability of change for each word in the dataset, it can then be evaluated against the ground truth of whether the word changed or not.
The area under the ROC curve corresponds to the probability that in a large random sample of (transmission failure, transmission success) pairs the model assigned a higher probability of change to the word that was actually changed vs. the word that was not changed.
The full ensemble model (AUC = .650, AIC = 23,504) outperforms simpler ones, for example one based only on word predictability (surprisal) under language models (AUC = .632; AIC = 23,900) and one based on invariant word properties (AUC = .583, AIC = 24,445).
This difference between areas under the respective AUC curves and the difference in AICs suggests that while both sets of predictors contribute to predictions, predictability under language models is a better predictor of transmission failure than invariant properties of words.

The previous ensemble model tests whether each model architecture makes an independent contribution in predicting transmission failures.
Another question is how the models directly compare to one another in their ability to predict transmission failures.
To this end, we fit a series of models with the same set of non-model-based predictors (e.g., position in sentence, age of acquisition, concreteness, and phonological neighborhood density) and use the per-word surprisal estimates from a single model.
We then compare the logistic models in terms of AIC and area under the ROC curve (AUC).
This analysis (Table \ref{tab:edit_prediction_comparison}) reveals that large LSTMs and smoothed 5-grams trained on a large collection of conversational text best predict transmission failures.
The Roark Parser has the least predictive power, though this may reflect that its training corpus (the Penn TreeBank) was at least a hundred times smaller compared to those used to fit the other language models.

\begin{table}
\caption{Comparison of individual language models for predicting transmission failures. Each model contains the same set of word properties, plus surprisal estimates from one language model. \label{tab:edit_prediction_comparison}}
\centering
\begin{tabular}{l|l|l}
Model              & AUC   & AIC   \\
\hline
OBWB / Big LM      & 0.639 & 23677 \\
DS / 5-gram        & 0.641 & 23687 \\
BNC / Trigram      & 0.638 & 23732 \\
BNC / Unigram      & 0.604 & 24095 \\
PTB / Roark Parser & 0.592 & 24370
\end{tabular}
\end{table}

%!!! good performance of the n-gram model, altogether

\section{Discussion}

% 1. overview of results
% #2. The role of preceding context
% #3. The Role of Abstract Structure
% #4. Word error model results
% #5. Artificial telephone results
% 6. General issues with the experiment
% 7. Next steps
% numbered items should get a subsection label

% 1. overview of results

In this work, we investigate several probabilistic generative language models (PGLMs) in their ability to capture people's linguistic expectations.
A serial reproduction task --- the game of Telephone, where participants reproduce recordings made by other participants in sequence---reveals participants' prior expectations in a naturalistic spoken word recognition task.
We find evidence that people use abstract representations of that context to inform their expectations, in line with the results of  \citet{fossumLevy2012} for reading.

\subsection*{The Role of Abstract Structure}

Before further interpretation of the results regarding the utility of abstract structure, we clarify the exact nature of our claims and note several methodological challenges that temper the strength and generality of these results.

First a top-level theoretical clarification is in order: Higher-level abstractions are thoroughly implicated in the processes of sentence comprehension and production (e.g., verb argument structure).
Our objective in this study, following \citet{frankBod2011}, is to investigate whether such abstractions may inform the lower-level process of word recognition.  

Second, we highlight an inherent brittleness in using a small cohort of probabilistic generative models to characterize human knowledge more generally: the encoded expectations reflect a complex interaction of architecture, training data, and fitting procedure.
Limited explanatory power for human performance may reflect any of one of the above aspects, or a complex interaction in between them.
Further, each language model is drawn from a larger space of possible language models, such that the generalizations we make about model architectures on the basis of a few examples may not be robust.

On the matter of fitting, the encoded expectations may reflect local minima or maxima for some models.
While this is not a concern with $n$-gram models fit with count-based methods, this problem increases in severity for language models which have large numbers of randomly-initialized parameters, especially neural network language models.
For these models, continued research focuses on how to improve the speed and robustness of the training procedure.
All models may suffer from the problem of overfitting \citep{gemanEtAl1992}, in which learned distributions reflect properties of the training data at the expense of their generality in extending to new data.
The relationship between corpus size and performance is modulated by architecture:
Large models trained on small datasets may be more prone to overfiting than small models trained on those same datasets.
More abstract structure may allow models to perform better on smaller datasets because they can fall back on expectations for higher-level abstractions in the absence of experience with specific words or sequences.
But smaller training corpora may also make it harder for models to arrive at useful abstractions in the course of fitting.

We took several steps to address these pitfalls regarding model performance.
First, we confirmed that all models exhibited characteristic levels of performance on standard test datasets, suggesting that the fitting procedures used here are consistent with previous benchmarks.
Though we note above the broader shortcomings of evaluating language models in terms of the perplexity on a held-out dataset, perplexity is nonetheless useful insofar as it allows us to check whether the fitting procedure yields models comparable or equivalent to those used in other studies.
Second, we treat the dataset on which each model is trained as a separate predictor in our analyses (binarized into those trained on the Penn TreeBank and those trained on large datasets known to yield highly performant models).
This helps isolate the contribution of the model architecture from the dataset on which the model was trained.
Third, we note that deficiencies in model fitting would most likely affect the more sophisticated neural network models or PCFG parsers, which would yield a null result rather than the positive one in favor of structure obtained here.

% broader conclusions
Another caveat is that the set of models investigated here represents only a small sample of possible models.
Sampling a larger set of models may reveal that the differences in fit to human behavior that we find for model architectures are not robust, or that other distinctions in model architecture are predictive of larger differences in performance.
We acknowledge this limitation of the current study, and note the possibility of evaluating a larger set of models---reflecting a larger space of architectures---in future work.

% Transition to interpretations considering this correct 
The regression analysis presented above yielded a statistically negative coefficient for \texttt{generation} $\times$ \texttt{abstract structure: used}, suggesting that models using abstract hierarchical structure better predict the changes made by people in the course of the transmission experiment.
However, we note that this is a relatively minor quantitative effect.
Several of the $n$-gram models, such as the DeepSpeech Kneser-Ney 5-gram and the BNC trigram model, exhibit a pattern of decrease in surprisal estimates that is very similar to models with considerably more sophisticated representations of abstract structure (BLLIP and Big LM).
This pattern of near-parity between certain large, higher-order $n$-gram models and more sophisticated generative models may be explained by a recent theoretical proposal regarding processing difficulty that posits graded use of detailed structural representations in people's linguistic expectations.
\textit{Noisy-context surprisal} suggests that while people may use structured, abstract representations of the preceding linguistic context, that such representations are imperfect and in particular tend to degrade the longer they are kept active in memory \citep{futrell2017noisy}.

\citet{futrell2017noisy} highlight the case of \textit{structural forgetting effects}, where people do not effectively use preceding grammatical structures to predict the remainder of the sentence.
For example, among the two utterances,

\begin{enumerate}
		\item *The apartment$_{1}$ that the maid${_2}$ who the cleaning service$_{3}$ had$_{3}$ sent over was$_{1}$ well-decorated.
		\item The apartment$_{1}$ that the maid$_2$ who the cleaning service$_{3}$ had$_{3}$ sent over was$_2$ cleaning every week was$_1$ well-decorated. 
\end{enumerate}

\noindent the first is grammatically ill-formed in that no verb corresponds with noun phrase with the head ``maid$_2$'', yet people give it consistently higher grammaticality ratings than the second sentence (originally presented in \citealp{gibsonThomas1999}).
This effect, they argue, arises from an ``information locality'' effect, whereby structural expectations---which should give infinitely more probability mass to the grammatically correct sentence than the incorrect one---are attenuated for material with longer intervening intervals (in terms of words or time).
Under this account, people have fleeting access to the representations (approximated with parse trees) that they form in sentence processing.
$n$-gram models can be thought of as an approximation of an abstract, structured generative model, but one with a particularly sharp memory decay function such that an extremely limited sample of the preceding context is used to predict the identity of the next word. 

The memory-based effect identified by  \citet{futrell2017noisy} may be further exacerbated in the current task by high levels of background noise, such that we observe relatively small advantages of abstract structural representations.
While imperfect memory imposes noise on the representation of context even under optimal acoustic conditions, this decay may be yet stronger if participants lack peaked estimates regarding what actually constitutes the preceding context.

%clarify the contribution of the regression model
\subsection*{An Error Model for In-context Spoken Word Recognition}

%main finding: residualized surprisal from the fancier models is predictive of transmission failures
Data regarding which words are successfully transmitted from speaker to listener  provide for an alternative method for evaluating how well these language models capture people's expectations.
Transmission errors are more likely when a word is unpredictable given the preceding context.
These models assign divergent probability distributions over continuations in many cases. 
For example, models with an abstract representation of syntax would assign different probabilities to the continuations ``is'' and ``are'' given the preceding context ``the quick brown fox and the lazy dog''.
A hierarchical model generally picks out a subject consisting of the coordinated NPs with heads ``fox'' and ``dogs,'' such that the plural copular ``are'' is a more likely continuation (less surprising) compared to ``is''. 
An $n$-gram model tracking a limited preceding lexical context, e.g., a trigram model that uses the only the two preceding words, would suggest that ``is'' is the more probable continuation given the two preceding words ``lazy dog''; under this model ``are'' is more surprising (*``the lazy dog are'').

The regression analysis above shows a predictive utility --- in the form of statistically significant beta coefficients---for trigram, 5-gram, PCFG (Roark), and LSTM (Big LM) estimates of in-context predictability---suggesting that even the most abstract representations of context are predictive of the collected transmission errors. 
Nonetheless, we find a continued role of word frequency, consistent with the predictions of the modeling work by \citet{goldwaterEtAl2010}.
A comparison of the individual models reveals that the LSTM (Big LM) and the DeepSpeech 5-gram model best predict the edits; we leave it to future work in the same format as \citet{goodkindBicknell2018} to test whether word predictability estimates from these models also excel at predicting reading times and other behavioral measures of processing difficulty.

The finding that the age of acquisition of a word is predictive of successful transmission---earlier-acquired words are more likely to be recognized than later-acquired ones---extends previous results from isolated visual word recognition \citep{morrisonEllis2000} into the auditory realm.
This is consistent with the hypothesis that such words enjoy a privileged status above and beyond their frequency, perhaps relating the problem of retrieving semantic representations \citep{austerweilEtAl2012} given the compositional structure of the lexicon \citep{vincentLamarreEtAl2016} in which lower frequency words are defined in reference to higher frequency ones.

A second notable finding is that once the relationship between word length and unigram surprisal is accounted for, longer words are \textit{more} likely to be recognized.
This can be interpreted as evidence that people are better able to recognize words that are more perceptually distinctive, in that words with more syllables have have \textit{fewer} perceptually similar competitors, above and beyond edit-distance-based measures of neighborhood density.
These words may also be easier to recover because there is more redundant information about their identity, such that listeners are more likely to be able to recover the intended word if some portion is corrupted in transmission. 
%!!! should do more here... cool data and cool model

%\subsection*{[[Serial Reproduction with Artificial Agents]]}

\subsection*{Limitations of Serial Reproduction}

A potential caveat to the generality of the results is that the expectations implicit in the utterances obtained from the Telephone game may be task-dependent, and of limited utility for characterizing linguistic expectations more generally.
For example, participants could infer that the task they are performing is qualitatively unlike ``normal'' language use, and make use of a different set of expectations such that the collected data is not representative of the distributions of interest for language processing.
It would be particularly concerning, for example, if the collected utterances demonstrated marked decreases in grammatical acceptability or semantic interpretability over the course of serial reproduction.
To evaluate whether participants produced grammatical and semantically-interpretable recordings for the duration of the experiment, authors S.M. and S.N. conducted a follow-up analysis in which they independently coded all utterances from generations 23-25 with binary judgments of grammaticality and semantic interpretability.
The latter category was used to distinguish sentences that are structurally well-formed but not interpretable, e.g., ``the bus and bus driver were opening the door.''
Utterances in this set were judged as 87\% and 88\% grammatical (Cohen's $\kappa$ = 0.94) and 78\% and 79\% semantically interpretable (Cohen's $\kappa$ = 0.89).
The results of this analysis suggest that participants largely maintain the grammaticality and semantic well-formedness of their responses through the course of the serial reproduction experiment, and suggest that participants are tapping into a similar set of expectations as ``normal'' language use.

Another important possibility is that participants could be modulating how they use preceding context in word recognition based on the level of noise in the experiment.
Because they may be unsure of the preceding context for a particular word, they may prefer shorter, less-structured representations of context in the current experiment, whereas they might rely on that context more heavily in a noise-free environment.
This basic logic of noise-modulated expectations is substantiated by the finding of \citet{lucePisoni1998} that participants' reliance on word frequency in isolated spoken word recognition increases as a function of the level of background noise.
Audio stimuli here are embedded in relatively high levels of noise, qualitatively dissimilar to the reading tasks of \citet{frankBod2011} and \citet{fossumLevy2012}.
We highlight the importance of characterizing variation in the use of linguistic expectations as a function of environmental noise as an important next step to explore with this paradigm.

%In this case, it would be unique interpreted as conclusive evidence against its use at lower levels of noise.
%However, it is worth noting that in the current case we find the \textit{opposite} pattern: even at high levels of noise, we find evidence that people make use of preceding context.

One possibility is that participants are providing responses according to the \textit{maximum a posteriori} (MAP) estimate, i.e., choosing the response with the highest posterior probability rather than sampling from possible responses according to their posterior probabilities.
If participants respond according to their MAP estimate of what was said by the previous speaker, the equivalence between the stationary distribution obtained in the limit and the participants' prior is no longer guaranteed.
Under certain conditions, the implications of participants choosing according to MAP are known, and the resulting stationary distribution is highly informative regarding the characteristics of the prior.
\citet{kirbyDowmanGriffiths2007} introduce a general-purpose method for interpolating between MAP and posterior sampling when hypotheses have a constant likelihood, and demonstrate that the ordering of hypotheses under the prior is preserved under MAP.
\citet{griffithsKalish2007} demonstrate the equivalence of MAP to expectation-maximization \citep{dempsterEtAl1977} in continuous hypothesis spaces, with a guarantee that the stationary distribution will be centered on the maximum of the prior, with variance increasing according to the rate at which  hypotheses change over sequential participants.
If the asymptotic normality assumption holds, this same equivalence between the stationary distribution and the prior should be expected here.

While the experimental paradigm presented here can potentially reveal participants' linguistic expectations, it is relatively costly from a sampling perspective.
Coverage in the dataset shared here is limited by the small number of initial sentences (reflecting a limited number of semantic contexts), relatively short chains, and relatively strong autocorrelation within those sampling chains.
While this method could in principle be used to collect a very large collection of utterances after convergence, from which a new probabilistic model of linguistic expectations could be estimated, a more data-efficient method may be to use the word-by-word responses to fine-tune the parameters of an existing model \citep{ziegler2019fine}.
In future work we will investigate whether this fine-tuning approach can effectively combine the broad coverage of language models learned from large text corpora with the specific behaviors of human participants in the task of in-context spoken word recognition.

Finally, we emphasize the limitations in the generality of these results with respect to individual variation among speakers of English, as well as variation across speakers of different languages.
This experiment makes the strong simplifying assumption that the process of serial reproduction yields samples from a single, unified set of linguistic expectations that are shared across all English-speaking participants for the purposes of word recognition.
Of course, linguistic expectations should vary significantly as a function of linguistic experience, and may vary between speakers for other reasons like population-level variability in working memory.
At a higher level, the expectations of English speakers are certainly not representative of the expectations of speakers in other languages.
Given the pronounced typological diversity of languages, expectations may take qualitatively different forms.
For example, listeners may rely less on sequential word order in languages with more flexible word order. 
Future work will be needed to characterize the ways and extent to which expectations vary across natural languages.

%\subsection*{Next steps}	
%	\subsubsection*{Vary the age of the participating %cohort}
%	\subsubsection*{Other uses of the corpus: much more data %for deletion; sociolinguistic variation}

\section{Conclusion}
In this study, we collected data on how utterances change in the course of a web-based game of ``Telephone.''
We use this data, which provides an alternative to corpus-based evaluation methods, to evaluate a range of broad-coverage probabilistic generative language models in their ability to characterize people's linguistic expectations for in-context spoken word recognition.
Addressing an outstanding empirical question in speech processing, we find that models that use an abstract  (i.e. hierarchical) representation of preceding context to inform expectations regarding upcoming words more strongly reflect the changes made by people.
This paradigm offers promise in helping to better understand humans' remarkable language processing abilities.

\section{Supplementary Material}

All audio recordings, metadata, and analyses can be accessed through a project-specific repository on the Open Science Foundation, osf.io/3hws2.

\bibliographystyle{apacite}
\bibliography{telephone_bibliography.bib}

\newpage
\section{Technical Appendix 1: Language Models and Training Procedure}

%!!!! something about how this is not exhaustive

\subsection{Language Models}

\subsubsection*{N-gram models} 

The simplest PGLMs that we evaluate here are $n$-gram models.
First used to model natural language by \citet{shannon1948}, $n$-gram models typically track forward transitional probabilities for words by conditioning on preceding words.
These models make the strong simplifying assumption that the sequence of words in an utterance or sentence are generated following a \textit{Markov chain}, in which the probability of each event (i.e., word) depends only on immediately preceding events. 
These models can condition on a larger or smaller preceding context: an $n$-gram model tracks conditional probabilities given $n$-1 preceding words: bigram models condition on just the preceding word, while trigram models condition on the two previous words.
The probability of an utterance is the product of the conditional probabilities of the constituent words, 

\vspace{-10mm}

\begin{align}
\label{eq:ngram}  
P(w_1,\ldots,w_m) = \prod_{i=1}^{m}{P(w_i| w_{i-(n-1)}, \ldots, w_{i-1})},
\end{align}

\noindent where $w_1,\ldots,w_m$ is an utterance comprised of $m$ words and $n$ is the order of the $n$-gram model. 
By convention, sentences (or utterances, more generally) are augmented with a start symbol (of probability 1) and an end symbol (the probability of which is tracked by the model, the same as any other word type).
These models track statistics over sequences of words, and do not include any explicit encoding of statistical expectations for higher-order abstractions like parts of speech (\eg nouns and verbs), super-ordinate grammatical categories (\eg NPs or VPs), or semantic roles (\eg agents or patients).
$n$-gram models may appear to encode expectations related to these abstractions because lexical statistics capture these regularities implicitly: a trigram model will assign almost all of the probability mass following ``near a'' to nouns and adjectives.
However, without support for abstract representations, a trigram model cannot assign higher probabilities to nouns as a class (i.e., nouns not observed following ``near a'') in predicting the next words in that context.
As such, comparing the probability of the serial transmission chains under the $n$-gram models against the probability under the other models---which all posit and track regularities at a level higher than words---allows us to evaluate evidence for the degree to which people may use representations of higher levels of linguistic structure to inform their expectations. 

A special case of the $n$-gram model of particular theoretical interest is the unigram, or 1-gram model, where the probability of a word is not conditioned on preceding context.
Unigram probability estimates thus largely reflect normalized word frequencies, but may assign nonzero probability mass to out-of-vocabulary words depending on the choice of smoothing technique, described below.
Comparing the predictions of the unigram model with those of higher-order $n$-gram models allows us to evaluate the degree to which participants' linguistic expectations reflect any amount of preceding context, separate from the treatment of abstraction. 

Besides the length of the context on which they condition, a second dimension of variation in the architecture of $n$-gram models is the choice of \textit{smoothing} technique, or how probability mass is re-allocated to unobserved word sequences during model fitting.
Using probabilities directly derived from counts, i.e., the maximum likelihood estimate, of an $n$-gram model is generally ill-advised because of sparsity: many word sequences observed in a new dataset may not have been observed in the dataset used to fit the corpus.
Unless unobserved sequences receive nonzero probability, sentences with zero-probability continuations will be evaluated as also having a probability of 0. 
This reallocation of probability mass among sequences often reflects simple assumptions about the statistical structure of languages.   
Smoothing over possible sequences complements the practice of assigning some proportion of the unigram probability mass to newly-encountered tokens. 

Here we use two smoothing schemes when higher-order $n$-gram models.
For larger datasets, we use modified Kneser-Ney smoothing \citep{chenGoodman1998}.
This smoothing scheme shifts probability mass to unobserved $n$-grams that are expected on the basis of the prevalence of constituent lower-order $n$-grams, e.g., assigning ``near San Antonio'' a small non-zero probability because ``near San'' (as in ``near San Francisco'') and ``San Antonio'' are both relatively probable bigrams.
For smaller datasets, we use Good-Turing smoothing \citep{galeSampson1995}, which shifts probability mass from $n$-grams seen once to ones that are not encountered in fitting.
We build several new $n$-gram models using the SRI Language Modeling Toolkit (SRILM, \citealp{stolcke2002}), and use a proprietary 5-gram model from the DeepSpeech project \citep{hannunEtAl2014} which was estimated using KenLM \citep{heafield2011}.
We continue here with the enumeration of language models; the linguistic datasets used to fit the models are treated in greater detail below.

\subsubsection*{Probabilistic Context-Free Grammars (PCFGs)}

The remainder of the probabilistic generative language models under consideration here use abstract representations above the level of the word to derive word-level expectations.
The first class of model, probabilistic context-free grammars (PCFGs), posit that utterances reflect a latent structure composed of abstract, hierarchical categories like nouns, verbs, noun phrases, and verb phrases.
The task of predicting the next word is thus not just dependent on the previously-observed words as in the $n$-gram models, but also depends on a listener's beliefs about the grammatical structure consistent with the previously-encountered words, and how that grammatical structure narrows the set of possible continuations.
For example, having heard words that likely form a verb phrase for a transitive verb (\eg ``bought a''), a listener might expect a noun phrase containing the object of that verb. 

Under a PCFG, each hypothesis about the hierarchical structure of an utterance is captured in terms of a parse tree, which describes how an utterance could be generated from a context-free grammar.
A context-free grammar is a linguistic formalism that describes sentences as a hierarchy of constituents, starting with a root ``sentence'' node.
This sentence node is composed of a tree of nonterminals: grammatical categories such as nouns, verbs, noun phrases, verb phrases, each of which in turn consists of other grammatical categories or words.
All terminals, or leaf nodes in the hierarchy, are (observable) words.
A CFG captures the intuition that the same (observable) linear sequence of words may reflect several possible (unobservable) interpretations of the relationship between constituents.
For example, ``the girl saw the boy with the telescope'' has two high probability interpretations dependingon the attachment point of the prepositional phrase --- whether ``with the telescope'' modifies \textit{boy} or how the girl \textit{saw}. 
While CFGs cannot capture certain human linguistic phenomena \citep{stabler2004}, probabilistic CFGs, or PCFGs, are commonly used as generative probabilistic models of language given their principled account of higher-order structure. 
Unlike typed dependency parsers (e.g. \citealp{deMarneffeEtAl2006}), which track typed pairwise relationships between words, PCFGs provide a probability distribution over hierarchical parses for an utterance. 

The parameters of PCFGs can be learned in an unsupervised fashion from linear word representations alone, or fit using corpora that have been annotated by linguists with gold-standard most-probable parses.
Because of their stronger performance, we focus here on the predictions of PCFGs derived from supervised training.
Given the size of the hypothesis space of possible grammars for a language, as well as parses for a sentence, PCFGs vary in the techniques that they use to find the highest probability parses.
Here we use two PCFGs, the \citet{roark2001} parser and the BLLIP parser \citep{charniakJohnson2005}, which approach the problem of inference in this large hypothesis space in very different ways. 

\subsubsection*{Roark Parser}

The Roark parser \citep{roark2001} is a widely used \textit{top-down} parser, meaning that it maintains and updates a set of candidate parse trees as it moves from left to right in an utterance.
In combination with other architectural decisions, this allows the Roark parser to calculate probabilities for each sequential word conditioned on the preceding words.
This incremental property makes it especially well-suited for modeling online sentence processing, where people update their interpretation of the utterance as they receive new acoustic data \citep{roarkEtAl2009}. 

\subsubsection*{BLLIP Parser}

Unlike the top-down approach of the Roark Parser, the BLLIP (or Johnson-Charniak) parser uses bottom-up inference in combination with a secondary scoring function on the yielded highest probability trees \citep{charniakJohnson2005}. 
The bottom-up approach means that the parser starts by positing grammatical categories directly above the level of words for the entire sentence, then iteratively identifies possible higher-level grammatical categories up to the sentence root.
As such, it cannot be used to produce probability estimates incrementally after each word.
The BLLIP parser uses bottom-up parsing to generate a set of high probability parses which are then re-ranked using a separate discriminative model with a researcher-specified set of ad-hoc features.
%[[more info without crashing the reader]]
Though bottom-up parsing violates the constraints of online auditory language processing in that it requires that the complete signal be received previous to parsing, we include the parser in analyses tracking utterance-level changes.  

For both PCFG language models, the probability of an utterance reflects a marginalization over the possible parse trees that generate that utterance.
Because of computational constraints, these parsers typically track a relatively small number of the highest probability parse trees (in this case, 50 for each model).
While this represents a truncation of the true set of possible parse trees, the first few (i.e., the first one to five) are overwhelmingly more probable than the remainder, and thus provide a reliable estimate of utterance probability.
In the case of the Roark parser, surprisal estimates for individual words can be derived by querying the probability of the next word given the restricted set of highest probability provisional parse trees \citep{roarkEtAl2009}.

\subsubsection*{Recurrent Neural Network Language Model (RNN LM)}

The expectations derived from PCFGs reflect hierarchical parses of sentences, but listeners could also develop expectations by noting abstract commonalities in the usage of words, for example that the words ``five'' and ``six'' tend to appear in very similar lexical contexts.
We include here two recurrent neural network models (RNNs) that can infer partially syntactic, partially semantic higher-level regularities in word usage, and can use these regularities in the service of prediction. 
As with PCFGs, at least some RNN architectures can track long distance dependencies \citep{linzenEtAl2016}.
While models lacking abstract representations of context could in principle capture long-distance dependencies (e.g., a 9-gram model), the combinatorial richness of language means evidence is far too sparse to infer these regularities when tracking sequences of words when treated as discrete symbols. 

Recurrent neural networks are able to capture and use these higher-level regularities because they use the state of the hidden layer of the network at the previous timestep, often referred to as a ``memory,'' in addition to newly-received data to make predictions.
This hidden layer from the previous timestep is a lossy---and thus generalizable---representation of the preceding context.
By projecting words in the preceding context into a lower-dimensional space, the observation of a specific word sponsors the activation of words with similar lexical distributions in the context, imbuing the model with robustness in prediction even if a particular sequence of words has not been seen.
Since early demonstrations of their utility for language prediction with small-scale, synthesized linguistic corpora \citep{elman1990}, an extensive literature has developed architectures to deal with web-scale natural language prediction tasks, particularly developing techniques to prevent overfitting given the massive number of parameters, and to manage their computational complexity \citep{mikolovEtAl2011}.

The first RNN we use here, ``RNN LM,'' was first described in \citet{mikolov2010}, while additional performance optimizations for training can be found in \citet{mikolovEtAl2011}.
We train a network with 40 hidden nodes that uses backpropagation through time (BPTT, \citealp{werbos1990}) for the two preceding timesteps.
The vocabulary is limited to the 9,999 highest-frequency word types, with the remainder of types assigned to an $<$unk$>$ type. 

%small-scale context-sensitive languages \citep{steijvers1996}
%[AD] \subsubsection*{Zaremba LSTM}
% Retrofitting Zaremba to output word-by-word probabilities is very painful

\subsubsection*{Big Language Model (Big LM)}

One of the major challenges with recurrent neural networks is the problem of \textit{vanishing gradients} where the error signal necessary to update connection weights in the network becomes too small to use standard gradient descent techniques.
This problem becomes especially prevalent when recurrent neural networks track longer histories of preceding events. 
The RNN language model described above, for example, can only update weights using the model state at two preceding timesteps.
One solution that has received significant attention is to use a special kind of node in the neural network's hidden layer, or \textit{long short-term memory} (LSTM) cells \citep{hochreiterSchmidhuber1997}.
While non-LSTM RNNs directly copy the previous state of the hidden layer at each timestep, the LSTM uses these special cells that regulate how information is propagated from one timestep to the next.
These cells have input, output, and forgetting ``gates'' that regulate how the cell's state changes, such that it can maintain state for longer intervals than a typical RNN. 
Such networks have been widely useful in sequence prediction tasks \citep{gersSchmidhuber2001}, with particular successes in language modeling \citep{sundermeyerEtAl2012, zarembaEtAl2014}. 
% http://harinisuresh.com/2016/10/09/lstms/
%http://colah.github.io/posts/2015-08-Understanding-LSTMs/

One challenge for RNNs, left unaddressed by the use of LSTMs, is the difficulty of scaling to larger vocabularies, a necessity for modeling large naturalistic language corpora.
Evaluating the network's loss function, which requires generating a probability distribution over all words, becomes prohibitively computationally expensive owing to the normalizing term in a softmax function.
This has lead researchers to limit vocabularies to sizes much smaller than those typical of $n$-gram models or PCFGs, often in the range of 5,000 to 30,000 word types. 

\citet{jozefowicz2015empirical} address this limitation in their ``Big LM'' architecture by representing words as points in a lower-dimensional continuous space using a convolutional neural network (CNN).
These models can represent words as embeddings (real-valued vectors) that capture perceptual similarity between words on the basis of shared structure, such as word lemmas or part of speech markings like \textit{-ing} \citep{lingEtAl2015}.
The LSTM is then used to make predictions in this lower dimensional space, reducing the number of computations in the softmax function from the size of the vocabulary to the dimensionality of the CNN-derived embeddings. 

Given the significant technical resources necessary to train such a model (24-36 graphics processing units) and that model hyperparameters have not been made publicly available, we use the model distributed by \citet{jozefowicz2015empirical}.
This model uses 4096-dimensional character embeddings to represent words, makes use of backpropagation through time for the previous 20 timesteps, and has two layers with 8192-dimensional recurrent state in each layer. 

\subsection*{Training Data and Model Fitting}

All of the above language models are fit or trained using corpus data represented as written text.
Ideally, we would evaluate all combinations of language models and corpora to identify how model performance reflects an interaction of model architecture and training data.
However computational limitations, licensing restrictions, and limited public availability of codebases make this goal unfeasible at present.
Instead, wherever possible we evaluate each language model trained on two datasets: one on a large corpus for which it is known to produce competitive or state-of-the-art results, and the other on the (publicly-available) contents of the Penn TreeBank \citep{marcusEtAl1993}. 

The Penn TreeBank is a standard training, validation, and test dataset in the public domain, consisting of about 930,000 word tokens in the training set.
For those models that take advantage of the validation dataset for setting hyperparameters, these models have additional access to 74,000 tokens.
All sentences are annotated with gold-standard parse trees, making this dataset appropriate for training the two supervised PCFG models.
While the thematic content and register of the Wall Street Journal are not representative of conversational English, all language models should be equally disadvantaged by this shortcoming. 

For the instance of each model architecture trained on a larger dataset, we use a variety of datasets with known strong performance in the psycholinguistics or NLP literature.
For $n$-gram models we use the British National Corpus (BNC; \citealp{BNC}), an approximately 200 million word dataset commonly used in psycholinguistics.
The BNC consists of material from newspapers, academic and popular books, college essays, and transcriptions of informal conversations.
The BLLIP parser is trained on the Google TreeBank \citep{biesEtAl2012} in addition to the Penn TreeBank. 
The Google TreeBank contains approximately 255,000 word tokens in 16,600 sentences with gold-standard parse trees, taken from blogs, newsgroups, emails, and other internet-based sources.
Big LM is trained on the One Billion Word Benchmark of \citet{chelbaEtAl2013}, a large collection of news articles.
The DeepSpeech project \citep{hannunEtAl2014} provides a smooth 5-gram model trained on a proprietary dataset including Librispeech \citep{panayotovEtAl2015} and Switchboard \citep{godfreyEtAl1992}. 
Only the Penn TreeBank was used in the cases of the Roark parser and the RNN LM. 

\subsection*{Computation of Lexical Surprisal}

Estimating surprisal --- the unexpectedness of a word, in the form of its negative log probability under a predictive model --- for each individual word token requires that a model produce conditional word probabilities.
These probabilities are straightforwardly accessible because $n$-gram models encode these continuation probabilities directly. 

The Roark parser produces an estimate of lexical continuation probability that marginalizes over the set of highest probability candidate parses given the words seen so far.
The bottom-up scoring procedure used by the BLLIP means that it assigns probabilities to parse trees that generate the whole sentence, rather than successive words.
For this reason, we analyze changes in sentence probability under the BLLIP, but do not use it as a predictor in analyses requiring word-level surprisal estimates. 

The RNN LM yields a vector of activations over the vocabulary which is translated with a softmax function into a probability distribution.
Big LM produces a probability distribution over continuations, though the softmax function is computed over the lower-dimensional representations of words.
In both cases the prediction by the neural network is conditioned on the state of the model at preceding timesteps.
We treat the probabilities obtained from models in the same way as the conditional probabilities from the $n$-gram models. 

\subsection{Computation of Sentence Probabilities}

For all models we omit the end of sentence marker in the computation of utterance probability.
We adopt this strategy because utterance-final punctuation was not collected in the Telephone game, and those models trained on datasets with punctuation assign very high surprisal to end-of-sentence markers without preceding punctuation.
In the case of $n$-gram models, sentence probability is simply the product of conditional word probabilities.

Evaluating the probability of a sentence for the two PCFG models, the Roark and BLLIP parsers, requires marginalizing over the set of possible parse trees.
For these two models, we sum over the probabilities of the top 50 parse trees yielded by each model.
While there may be many more trees that would yield the same string of words, the first few (1-5) highest ranking parses typically account for nearly all of the aggregate probability mass for that utterance.

As first noted by \citet{bartlett1932}, messages decrease in length over the course of serial reproduction.
Shorter utterances are necessarily more probable, in that they are comprised of a smaller number of events. To investigate the effects on utterance probability separate from utterance length, we normalize the negative log probability by the number of words in each utterance.
This measure, which we call ``average per-word surprisal'' has an intuitive interpretation in terms of bits of information.
However, this measure may disguise an important difference between models in the interpretation of utterance probabilities among the PCFGs.
For the non-incremental models (\eg BLLIP), the probability reflects the set of parse trees consistent with the complete utterance.
By contrast, the set of parse trees changes after observing each word under the incremental models.
Garden path sentences provide a case where the latter class of models might assign high probability to the initial sequence of words ``the horse raced past the barn'', then a low probability to the continuation ``fell,'' which is likely to be low probability under the set of parses in the beam. 
By contrast, a bottom-up, non-incremental model would assign a low probability to all expansions necessary to generate this sentence, in that it only considers the subset of parses compatible withe the full utterance.

\section{Technical Appendix 2: Experiment Interface}

Participants are directed from Mechanical Turk to an externally-hosted version of the Telephone Game.
The app guides the participant through adjusting their speakers and microphone, after which participants proceed through a sequence of four practice trials.
The first practice trial tests whether the participant's speakers or headphones are working properly by having them transcribe a ten-word sentence.
The second practice trial tests whether a participant's microphone is working by having them listen to a recording of a sentence and repeating its content.
At the beginning of this trial, the participant grants permission in the web browser for the use of their microphone.
Participants are prompted to repeat the second practice trial until their recording is similar to the gold-standard transcription of that sentence, as evaluated by a normalized Levenshtein-Damerau distance of .2 between the gold-standard transcription and the output of an automatic speech recognition system run on their recording (described in greater detail below).

After completing the practice trials, the participant begins a series of 47 test trials, including 40 target trials and 7 randomly interspersed fillers. 
The 40 target trials consist of either sentences from a pool of initial recordings (described below) or the most recently-contributed recording of another participant.
The seven filler utterances are the complement of the four filler utterances used for the practice trials, and are drawn from an inventory of nine audio recordings with similar properties to the initial stimuli recordings, plus two standard test sentences from the TIMIT corpus meant to elicit variation in American English dialects \citep{garofolo1993}.
If the participant flags the utterance that they heard as inappropriate in (2) or their own recording as compromised in some way (4),the trial ends early, and they progress to the next trial.
Otherwise, if the recording passes the set of automated tests, then the state of the experiment is updated after each target trial to point to that newest recording as the appropriate stimulus for the next participant.

An example screenshot is presented in Fig. \ref{fig:telephoneInterface}.
We now present in greater detail the five steps an experimental trial and the automated filters outlined above.

\subsubsection*{1. Listening to a recording}

The participant clicks a button to start the trial, which triggers a three-minute audio timer that remains visible to the user through Step 5, below. 
This three minute timer ensures that the web experiment (given the dependencies between participants) is not blocked by an inactive participant; if time expires, another participant receives this stimulus.
Participants receive friendly feedback to encourage them to complete trials more quickly if this timer expires.
Besides the button to start playing the audio recording, there are no controls to pause, repeat, or move location in the recording.
The audio is presented embedded in noise recorded from a coffee shop (average SNR$_{dB}$ across recordings = -6.8).
Pilot experiments established that this naturalistic source of noise introduced a high rate of edits without introducing significant participant attrition, as was observed with similar amplitude white noise.
Each utterance has between 500 and 1000 ms of noise-only padding preceding and following the speech content.

\subsubsection*{2. Flagging the upstream recording}

Though the participant cannot pause or move within the recording that they hear, they may flag the recording at any time. 
If the participant chooses to flag a recording, they are then asked to choose one of a set of provided reasons: Contains speech errors, Speech starts or stops abruptly, Contains obscenities, and Other.
Upon choosing Other, the participant could provide a free-form text response.
The trial ends early if any of these options are chosen.

\subsubsection*{3. Recording best guess of what was heard}

If the participant does not flag the audio recording, they are immediately prompted to record their best estimate of what was said, with the specific prompt of ``Repeat the sentence you just heard as best you can.''
After clicking the Record button, the waveform for the recording is drawn in real time.
The participant may listen to and re-record their response as many times as desired.
If a recording is more than eight seconds long, the participant is prompted to record again.
When finished, the participant clicks ``Submit.''
Because the HTML5 audio recording specification leaves the choice of sampling rate and bit depth to the client, all recordings are normalized upon receipt by the server to 16 kHz, 16 bit PCM WAV files.
The acoustic properties of the recording environments vary between participants in that each participant records their responses in an uncontrolled environment.

\subsubsection*{4. Self-flagging the new recording}

After submitting a recording, a participant is given the opportunity to ``self-flag'' it in case of a speech error or other problems such as unexpected background noise.
If the participant chooses to self-flag the recording, they are prompted to provide a reason, with the same set of candidate reasons as the upstream flagging procedure in Step 2, above.
If a participant self-flags, the trial ends early.

\subsubsection*{5. Transcribing the new recording}
 
Finally, if the participant has submitted a recording and has attested that it is of good quality, they are then prompted to provide a written transcription thereof.
The inclusion of punctuation or of a misspelled word (as determined with the Linux utility \texttt{Aspell}) returns an error message to the participant, who may then edit the transcription and resubmit.
After submitting, the trial ends. 
The participant is then provided with a button entitled ``Click to play next audio recording''; because there is no timer on this screen, they may pause between each trial.

\subsubsection*{Automated filters}

After the participant submits the written transcription, they advance to the start button for the next trial. 
The web application asynchronously applies a set of automated filters to the audio file and the transcription on the server.
We employ these filters to determine if a recording is of sufficient quality to be used as input to other downstream participants.
In principle, these filters could be implemented with human participants, but using automated tools allows us to direct participants' effort (and commensurate compensation) towards data collection. 
Of note from a design and data analysis perspective, these filters \textit{must} be applied at the time of collection: because future participants hear responses from earlier participants, responses must be filtered in real time to maintain the continuity and integrity of the recording chains.
These filters include: 

\begin{itemize}
\item Is the file silent?  
\item Is the transcription provided by the new participant between 20\% longer and 20\% shorter (in terms of the number of non-space characters) than the transcription of the input sentence they received?
If utterances become too short, they become difficult to characterize with language models.
\item Is the transcription provided by the new user more than 2 words longer or 2 words shorter than the transcription of the input sentence? This follows the same logic as above.
\end{itemize}

A second set of tests pertains to the audio quality and the intelligibility of the recording. 
For this we use an automatic speech recognition system to generate a transcription of the newly-recorded audio file. 
Specifically, we use one of the most advanced publicly-available automatic speech recognition systems, DeepSpeech \citep{hannunEtAl2014}, to check:

\begin{itemize}
\item Is the DeepSpeech-generated transcription of the participant's audio file similar to the transcription they provided? This guards against the possibility that a user would provide an  acceptable transcription, but an unrelated audio file (e.g., one filled with obscenities).
\item Is the DeepSpeech-generated transcription of the participant's audio file similar to the transcription provided by the upstream participant?
This prevents the introduction of material like ``I didn't hear the last sentence.''
\end{itemize}

The above checks are operationalized by testing whether the normalized Levenshtein-Damerau distance \citep{navarro2001} between the strings in question is less than a precomputed threshold.
We use the threshold of .58, arrived at by computing the normalized Levenshtein distance between each sentence in a large corpus and a number of candidate transcriptions, including the correct one and several unrelated foils.
On this test corpus with highly dissimilar sentences, this threshold yields a negligible false positive rate; in practice, this threshold is extremely permissive (allows changes) and flags only highly deviant recordings.  
The length filter was motivated by the finding in pilot work that utterances decreased in length very rapidly without such constraints, yielding many 1-3 word utterances of questionable linguistic status.

If a recording fails any of the above tests, the participant receives feedback at the end of the following trial.
This asynchronous evaluation allows for efficient speech recognition (which is computationally costly and requires the use of a graphics processing unit on the server) and prevents participants from having to wait for the web application to recognize and validate their responses.
Utterances that are rejected by any of these filters are retained in that they are potentially relevant to a number of research questions outside of the scope of the current study.
If a recording passes the above tests, the recording is combined with a randomly selected interval of cafe background noise (with the same acoustic properties as above) so that it may be used as a stimulus for later participants.
The implications of flagging the upstream stimulus, self-flagging, and automated filtering for the stimuli heard by later participants are described below in relation to the process of serial reproduction.

After submitting a response for the final stimulus, participants are prompted to take an optional demographic survey detailing age, gender, level of education, current geographical location, proficiency with English, and information regarding previous residence for the purpose of dialectal analyses (outside of the scope of this work).

\section{Technical Appendix 3: Experiment Transmission Structure}

The succession of stimuli and responses (the latter constituting stimuli for later participants) can be conceived of in terms of a directed acyclic graph, or DAG. 
Considering the succession of recordings for each sentence as a graph---a collection of nodes representing recordings and edges representing the history among participants---allows us to concisely represent the data collected in the course of the experiment and to operationalize the logic for both automated and participant-based flagging of recordings. 
% [AD] FIG: Example DAG for a single sentence

Per the specification of the interface above, a recording takes one of five possible states: accepted by fiat because it is an initial stimulus (``protected''), provisionally accepted (``accepted''), flagged by a downstream participant (``downstream-flagged''), flagged by the participant who recorded the sentence (``self-flagged''), or flagged by one of several automated methods (``auto-flagged'').
When a participant starts a test trial indexed by a particular initial sentence, they are provided with either the most recent accepted recording or the initial recording itself (if no previous recordings have been accepted).
In other words, if a participant flags the input recording they heard, the following participant will then hear the previously accepted recording in the graph for that sentence.
The sequence of recordings appropriate for analysis, or \textit{recording chain}, is then the initial sentence recording and the subsequent succession of accepted recordings.
Flagged and self-flagged recordings comprise the complement of the nodes in the graph.
We follow the convention established in the iterated learning literature of referring to the sequential position of a recording within a chain as its \textit{generation}, though note that a participant in this setup may contribute recordings for different stimuli at different generations, unlike most iterated learning experiments. 

The process by which participants are assigned to stimuli can then be considered in terms of \textit{threading}.  
Each participant must provide recordings for each of the 40 test sentences.
Because of the need for strictly successive recordings, only one participant at a time may listen to and record a response to the same recording.
We implement these constraints by implementing a \textit{mutex}, a data structure that limits concurrent access to a resource, in this case the recording chains.
In combination with multiple independent chains for each stimulus, this setup allows participants to listen and record sentences continuously while maintaining a strictly sequential relationship between the contributed recordings.
The three-minute timer on each trial means that the controller (the web app) will re-assign a stimulus to another participant if a response is not recorded within three minutes.
This setup also means that the order in which a participant contributes to each of the recording chains is randomized across participants.

These dynamics mean that any recording may be flagged and removed for the duration of the experiment, even if another participant records a downstream utterance.
For example, it is possible that sentence $s_1$ recorded by participant $p_1$ is provisionally accepted, and that a succeeding participant $p_2$ records a downstream repetition, $s_2$. 
If, however, $p_3$ flags $s_2$, then $p_4$ will hear $s_1$, and may in principle flag that recording.
We find that such cases of retroactive flagging (and loss of analyzable data) are relatively rare, but that this mechanism provides an automated and scalable method to produce chains of interpretable utterances appropriate for analysis.
This architecture also means that if two participants $p_1$ and $p_2$ are progressing through the experiment at the same time, then $p_1$ may provide the stimulus recording heard by $p_2$ for some sentences and $p_2$ may provide the stimulus recording heard by $p_1$ in others.

\section{Technical Appendix 4: Analysis of Flagged Recordings}

15.3\% of participant-contributed audio recordings were flagged by downstream participants as too noisy, ungrammatical, or otherwise unintelligible.
Specifically, 4.5\% were classified by participants as ``Contains speech errors'' (ungrammatical), 3.5\% as ``Speech stops or starts abruptly / speech sounds cut off,'' and the remainder were flagged for a variety of user-specified reasons regarding audio quality, e.g., ``too hard to hear,'' ``drowned out,'' ``can't hear it.''  

\begin{figure}
\begin{center}
\includegraphics[width=4.5in]{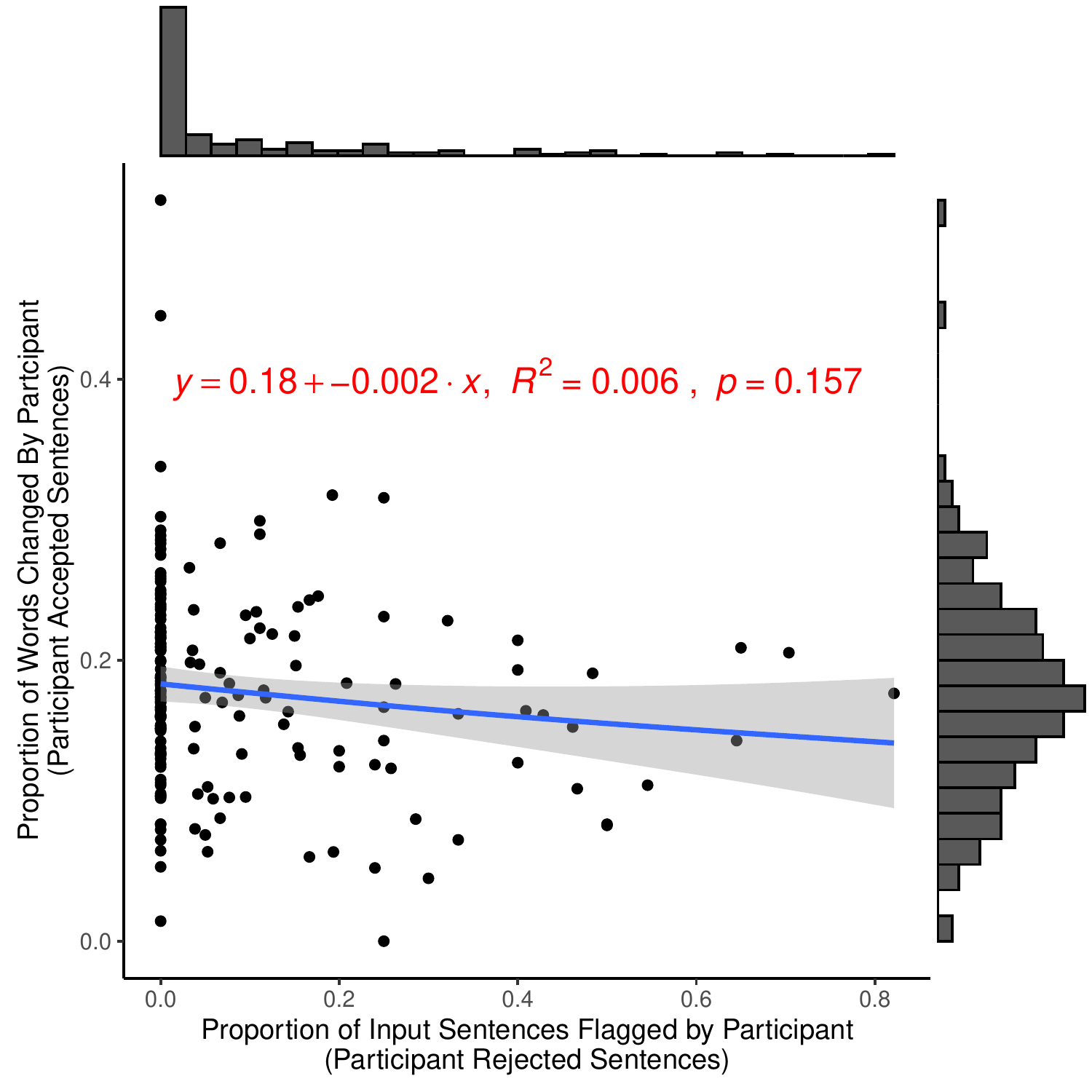}
\end{center}
\vspace{-5mm}
\caption{\label{fig:wordChangesVsFlags} 
Relationship between participant flag rate (i.e., number of recordings rejected) and participant edit rate (proportion of words changed among accepted recordings). 
Most participants change around 1 in 5 words that they hear, and flag none of the recordings that they hear. 
The non-significant correlation suggests that there is no relationship between flag rate and edit rate by participant. 
}
\end{figure}

One possibility that would complicate interpretation of the results presented here would be if participants who flag a high proportion of recordings behave differently (i.e., make fewer or more edits to what they hear) than those who flag a low proportion of recordings.
This could arise if some characteristic of the participant affected both the flag rate and the edit rate.  
In principle, a participant's robustness to noise (e.g., robustness in speech recognition related to age or language experience) could affect both: a participant with better hearing may flag few sentences and produce a high proportion of matches.
However, we do not find a statistically significant correlation between the proportion of words changed (deleted or substituted) and the proportion of input sentences flagged by a participant (Fig \ref{fig:wordChangesVsFlags}, $R^2$ = .006, $p$ = .157). 
This suggests that participants who flag many recordings do not behave differently than those who flag few recordings.

%!!! also refer to the edit rate on the flag rate

\section{Technical Appendix 5: Similarity of Language Models}

In the main text we present the pairwise similarity of all models trained on the Penn TreeBank in order to isolate the contribution of model architecture (vs. training data) to surprisal estimates.
Here we present the same measure of pairwise similarity, but for all 11 models.
This reveals that $n$-gram models trained on the PTB constitute one well-defined cluster separate from all other models. 
Another cluster emerges among Kneser-Ney trigrams and 5-grams trained on large datasets and the Big LSTM trained on the One Billion Word Benchmark.
The BLLIP PCFGS constitute a 2-model cluster. 
The Roark Parser, the RNNLM and unigram probabilities from the BNC constitute a separate set from the BLLIP PCFGs, but pattern more closely with them than the large models or the PTB-trained $n$-gram models.

\begin{figure}[t]
\begin{center}
    \includegraphics[width=\textwidth]{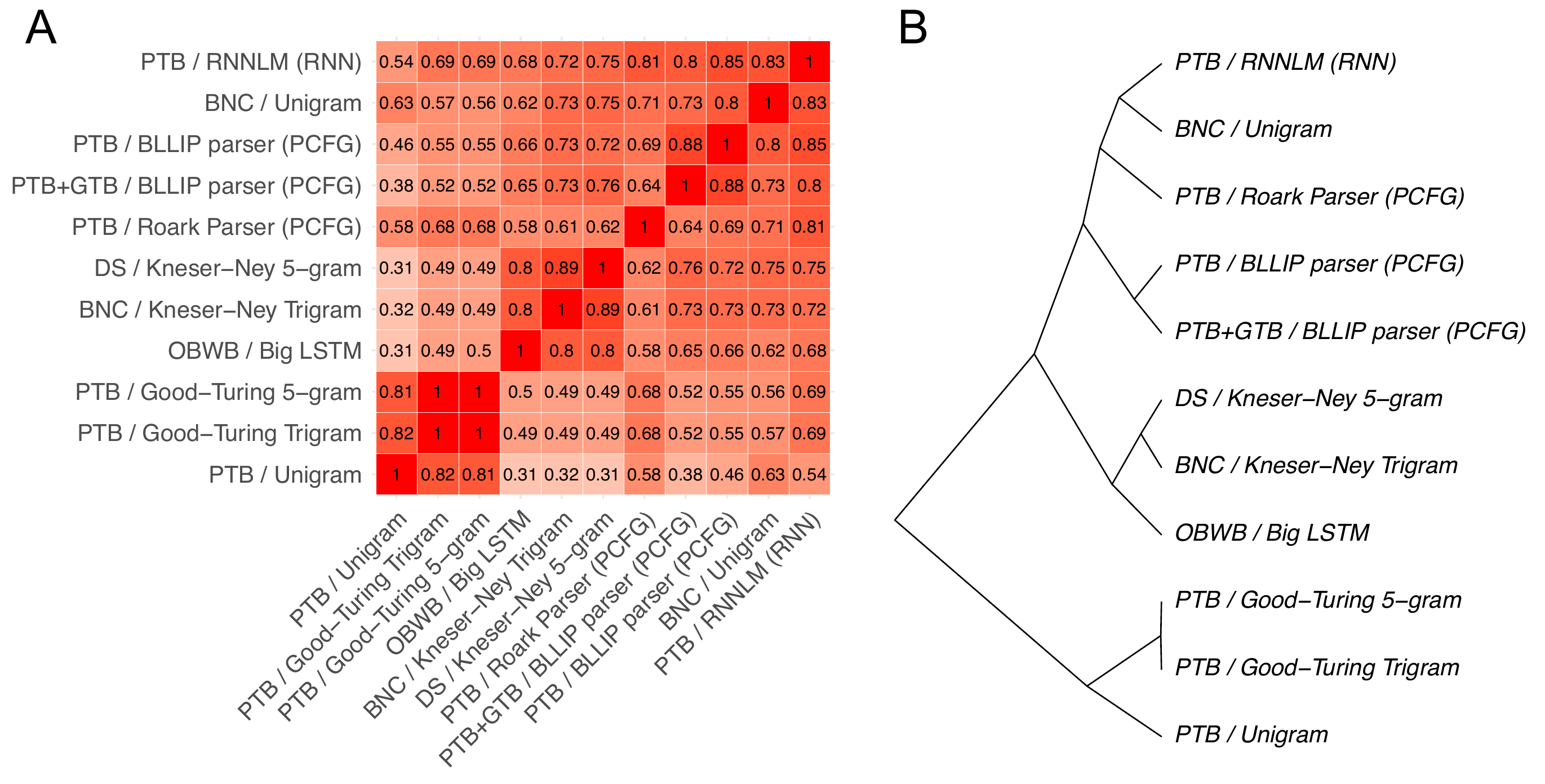}    
  \caption{Model similarity across 11 language models. \textit{A.} Spearman's rank correlation for pairwise combinations of sentence probabilities across the 3,193 sentences from the experiment across all models. \textit{B.} The resulting dendrogram, derived from the rank correlations using Ward's method.}
  \label{fig:lm_similarity}
 \end{center}
\end{figure}

\end{document}